\newcommand{\parshrinky}{\vspace{-4mm}}
\newcommand{\parskiny}{\vspace{-2mm}}
\newcommand{\seckiny}{\vspace{-2mm}}
\newcommand{\figskiny}{\vspace{-4mm}}
\newcommand{\figsmall}{\vspace{-1mm}}
\newcommand{\figshrinky}{\vspace{-2mm}}

\newcommand{\done}[1]{}%{}{{\tiny \textcolor{green}{DONE: #1}}}

\documentclass[10pt,twocolumn,letterpaper]{article}

\usepackage{cvpr}
\usepackage{times}
\usepackage{epsfig}
\usepackage{graphicx}
\usepackage{amsmath}
\usepackage{amssymb}
\usepackage[font=small]{caption} 
\usepackage{subcaption} 
\DeclareMathOperator*{\argmax}{argmax}
\usepackage[breaklinks=true,bookmarks=false]{hyperref}

\cvprfinalcopy % *** Uncomment this line for the final submission

\def\cvprPaperID{1518} % *** Enter the CVPR Paper ID here

\ifcvprfinal\fi
\begin{document}

%%%%%%%%% TITLE
\title{An Active Search Strategy for Efficient Object Class Detection}
{\small
\author{Abel Gonzalez-Garcia\\
%University of Edinburgh\\
{\tt\small a.gonzalez-garcia@sms.ed.ac.uk}
\and
Alexander Vezhnevets\\
%University of Edinburgh\\
{\tt\small avezhnev@inf.ed.ac.uk}
\and
Vittorio Ferrari\\
%University of Edinburgh\\
{\tt\small vferrari@staffmail.ed.ac.uk}
\and
University of Edinburgh\\
}
}

\maketitle
\thispagestyle{empty}

%%%%%%%%% ABStrACT
\begin{abstract}
Object class detectors typically apply a window classifier to all the windows in a large set, either in a sliding window manner or using object proposals.
In this paper, we develop an active search strategy that sequentially chooses the next window to evaluate based on all the information gathered before. 
This results in a substantial reduction in the number of classifier evaluations and in a more elegant approach in general. 
Our search strategy is guided by two forces.
First, we exploit context as the statistical relation between the appearance of a window and its location relative to the object, as observed in the training set.
This enables to jump across distant regions in the image (e.g. observing a sky region suggests that cars might be far below) and is done efficiently in a Random Forest framework.
Second, we exploit the score of the classifier to attract the search to promising areas surrounding a highly scored window, and to keep away from areas near low scored ones.
Our search strategy can be applied on top of any classifier as it treats it as a black-box.
In experiments with R-CNN on the challenging SUN2012 dataset, our method matches the detection accuracy of evaluating all windows independently, while evaluating $9\times$ fewer windows.
%It even outperforms it when evaluating $3\times$ fewer windows.
\end{abstract}

%%%%%%%%% BODY TEXT
\seckiny
\section{Introduction}
\figsmall
Given an image, the goal of object class detection is to place a bounding-box around every instance of a given object class.
Modern object detectors~\cite{cinbis13iccv,Dalal05:thomas, felzenszwalb10pami, girshick14cvpr,  harzallah09iccv, MalisiewiczICCV11, uijlings13ijcv,  wang13iccv} partition the image into a set of windows and then score each window with a classifier to determine whether it contains an instance of the object class. 
%This task is commonly tackled by training a window classifier that detects whether a window contains an object or not, and then evaluating a fixed, pre-defined set of locations in the image.
%The output of the detector is all the windows corresponding to the set of local maxima of the classifier scores, which are considered potential locations for the objects.
The detector finally outputs the windows with the locally highest scores. % NMS

In the classical sliding window approach~\cite{Dalal05:thomas, felzenszwalb10pami, harzallah09iccv, MalisiewiczICCV11}, the window set is very large, containing hundred of thousands of windows on a regular grid at multiple scales.
This approach is prohibitively expensive for slow, powerful window classifiers which are state-of-the-art nowadays~\cite{cinbis13iccv,donahue13decaf,girshick14cvpr,uijlings13ijcv,wang13iccv} such as Convolutional Neural Networks (CNN)~\cite{girshick14cvpr}.
For this reason, these detectors are based instead on object proposal
generators~\cite{alexe12pami,manen13iccv, uijlings13ijcv},
which provide a smaller set of a few thousand windows likely to cover all objects.
Hence, this reduces the number of window classifier evaluations required.
However, in both approaches the window classifier evaluates {\em all} windows in the set, effectively assuming that they are independent. 
%Nonetheless, windows are not independent.
%Even if a window is not on the object, it might contain information about the object's location.

%Description Object detection and why it is important. Description of the sliding
%windows/object proposals techniques. Sliding windows brute-force, unnatural and it wastes computation,
%not allowing the use of expensive classifiers.  Although proposals is better,
%it is still wasteful and rigid, looking at pre-defined set of locations, unlike
%human visual search.\\

\begin{figure}
\begin{center}
  \includegraphics[width=0.37\textwidth]{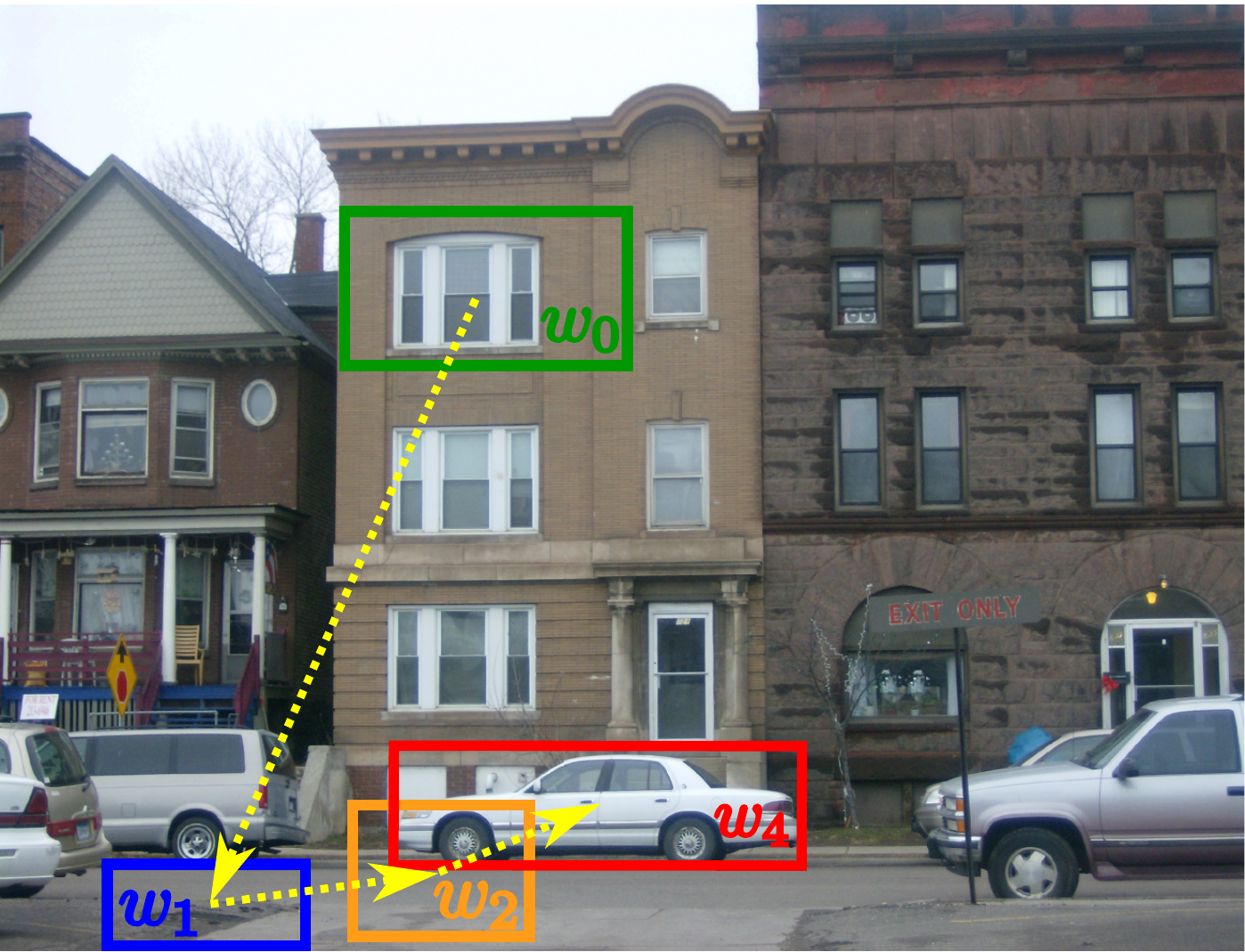}
\end{center}
\figskiny
  \caption{\small \bf Example of a car search using our method. \figskiny \figshrinky}
  \label{fig:searchEx}
\end{figure}

In this work, we propose an \emph{active search strategy} that sequentially chooses the next window to evaluate based on previously observed windows, rather than going through the whole window set in an arbitrary order. 
%
%Making sequential observations may provide information about the location of the object. 
%A window \emph{observation} not only provides information regarding the presence of the object in its location, but also about its surroundings and even distant areas of the image. 
Observing a window not only provides information about the presence of the object in that particular window, but also about its surroundings and even distant areas of the image. 
Our search method extracts this information and integrates it into the search, effectively guiding future observations to interesting areas, likely to contain objects. 
Thereby, our method explores the window space in an intelligent fashion, where future observations depend on all the information gathered so far.
This results in a more natural and elegant way of searching, avoiding wasteful computation in uninteresting areas and focusing on the promising ones.
As a consequence, our method is able to find the objects while evaluating much fewer windows, typically only a few hundreds (sec.~\ref{sec:results}).
%
%What is the paper about?\\
%
%Development of active search strategies that instead of looking at a pre-defined
%and fixed set of locations (either sliding windows or object proposals),
%sequentially explore the window space in an intelligent fashion, where the next
%observation depends on all the information gathered before.\\

We use two guiding forces in our method: context and window classifier score.
Context exploits the statistical relation between the appearance and location of a window and its location relative to the objects, as observed in the training set.
For example, the method can learn that cars tend to be on roads below the sky. 
Therefore, observing a window in the sky in a test image suggests the car is likely to be far below, whereas a window on the road suggests making a smaller horizontal move.
We learn the context force, in a Random Forest framework that provides great computational efficiency as well as accurate results.
The classifier score of an observed window provides information about the score of nearby windows, due to the smoothness of the classifier function.
It guides the search to areas where we have observed a window with high score, while pushing away from windows with low score.
Observing a window with part of a car, for example, will attract the search to its surroundings.

Our method effectively combines these two forces.
Fig.~\ref{fig:searchEx} shows the intuition of our method on detecting cars.
It starts at window $w_0$ and it moves away immediately, since $w_0$ contains a piece of building, not a car. 
Context determines the direction of the move, as cars tend to be on streets below buildings. 
Hence, the next visited location is on the road. 
After observing $w_1$, the method continues searching along the road, as indicated by context. 
For window $w_2$, however, the score of the classifier is rather high, as it contains a piece of car.
Therefore, the search focuses around this area until it finds a tight window on the car.
%In these last steps, context is still helpful, as our training database of windows lso contains windows on the objects. 

%Why is this interesting?\\
%Because it is a more natural and elegant way of performing object detection,
%that avoids exploring areas of the image which are not interesing, while
%focusing on those with higher potential to contain the object. It speeds-up
%the object detection process and also paves the way for more expensive, but more
%accurate classifier, since the number of evaluations is reduced.\\
%

Experiments on the challenging SUN2012 dataset~\cite{xiao10cvpr} and PASCAL VOC10~\cite{Everingham10} demonstrate that our method explores the image in an intelligent way, effectively detecting objects in only a few hundred iterations.
As window classifiers we use the state-of-the-art R-CNN~\cite{girshick14cvpr} and the popular UvA Bag-of-Words model of~\cite{uijlings13ijcv}, both on top of object proposals~\cite{uijlings13ijcv}.
For R-CNN on SUN2012, our search strategy matches the detection accuracy of evaluating all proposals independently, while evaluating $9\times$ fewer proposals. % at this point our mAP is less than 1% below eval all proposals (at this scale, fair concept of 'matching')
As our method adds little overhead, this translates into an actual wall-clock speedup.
When computing CNN features on the CPU~\cite{jia13caffe}, the processing time for one test image reduces from 320s to 36s ($9\times$ speed-up). When using a GPU, it reduces from 14.4s to 2.5s ($6\times$ speed-up). 
Hence, our method opens the door to using expensive classifiers by considerably reducing the number of evaluations while adding little overhead.
For the UvA window classifier, our search strategy only needs 35 % at this point our mAP is actually a tiny bit better, so 'exact matching'
proposals to match the performance of evaluating all of them (a reduction of $85\times$). By letting the search run for longer, we even {\em improve} accuracy while evaluating $30\times$ fewer proposals, as it avoids evaluating some cluttered image areas that lead to false-positives.

%\todo{paper plan?}

\seckiny
\section{Related Work}
\figsmall
\paragraph{Object proposals.}
Recent, highly accurate window classifiers like high-dimensional Bag-of-Words~\cite{uijlings13ijcv} or CNN~\cite{donahue13decaf, girshick14cvpr, krizhevsky12nips} are too expensive to evaluate in a sliding window fashion.
For this reason, recent detectors~\cite{cinbis13iccv,girshick14cvpr,uijlings13ijcv,wang13iccv} evaluate only a few thousands windows produced by object proposals generators~\cite{alexe12pami,manen13iccv,uijlings13ijcv}.
The state-of-the-art detector~\cite{girshick14cvpr} follows this approach, using CNN features~\cite{krizhevsky12nips} with Selective Search proposals~\cite{uijlings13ijcv}.
Although proposals already reduce the number of window classifier evaluations, our work brings even further reductions.

\parskiny
\paragraph{Improving sliding window.}
Some works reduce the number of window classifier evaluations.
Lampert et al.~\cite{Lampert08cvpr} use a branch-and-bound scheme to efficiently find the maximum of the classifier over all windows.
However, it is limited to classifiers for which tight bounds on the highest score in a subset of windows can be derived. 
Lehman et al.~\cite{lehmann2011bmvc} extend \cite{Lampert08cvpr} to some more classifiers. 
Sznitman et al.~\cite{sznitman2010pami} avoid exhaustive evaluation for face detection by using a hierarchical model and pruning heuristics.

An alternative approach is to reduce the cost of evaluating the classifier on a window.
For example,~\cite{harzallah09iccv,Vedaldi09} first run a linear classifier over all windows and then evaluate a complex non-linear kernel only on a few highly scored windows.
Several techniques are specific to certain types of window classifiers and
achieve a speedup by exploiting their internal structure (e.g.
DPM~\cite{felzenszwalb10cvpr_b,pedersoli11cvpr,zhu14eccv},
CNN-based~\cite{he14eccv}, additive scoring functions~\cite{wu13iccv}, cascaded
boosting on Haar features~\cite{saberian14jmlr,Viola01}.
Our work instead can be used with any window classifier as it treats it as a black-box.

%\todo{what about karayev14cvpr?? AG: It is definitely image classification (they call it object recognition), although the reviewer presented it as object detection, hence the confussion. Not sure where to put it because they don't do sequential fixations, like the other works under the image classification category.}
A few works develop techniques that make sequential fixations inspired by human perception for tracking in video~\cite{bazzani11icml}, image classification~\cite{denil12nc,larochelle10nips,mnih14nips} and face detection~\cite{butko09cvpr,tang14nips}.
% butko09cvpr: only faces, also foveated; only real object detection work
% bazzani11icml: tracking in video, not detection, and only MNIST digits and easy faces
% both of those state they are motivated by human perception
% larochelle10nips: synthetic datasets, MNIST digits, whole-image facial expression recognition; classification tasks! foveated, so also inspited by human perception; not detection! No real search.
% mnih14nips: recurrent model, MNIST digits, non-vision 'catch game' data; classification tasks
% tang14nips: caltech4 faces, eval localization a little bit, non-standard; not full image search, just start 100-pixels away from true face localition, crappy WSL experiments
% denil12nc: extension of larochelle10nips, same properties
% nobody's got a context force
However, they only use the score of a (foveated) window classifier, not exploiting the valuable information given by context.
Moreover, they experiment on simple datasets, far less challenging than SUN2012~\cite{xiao10cvpr} (MNIST digits, faces).
% well, the face dataset of butko09cvpr appears difficult, but the one in larochelle10nips is trivial, it's for whole-image classification; so approx here

\begin{figure*}
  \begin{center}
  \includegraphics[width=\textwidth]{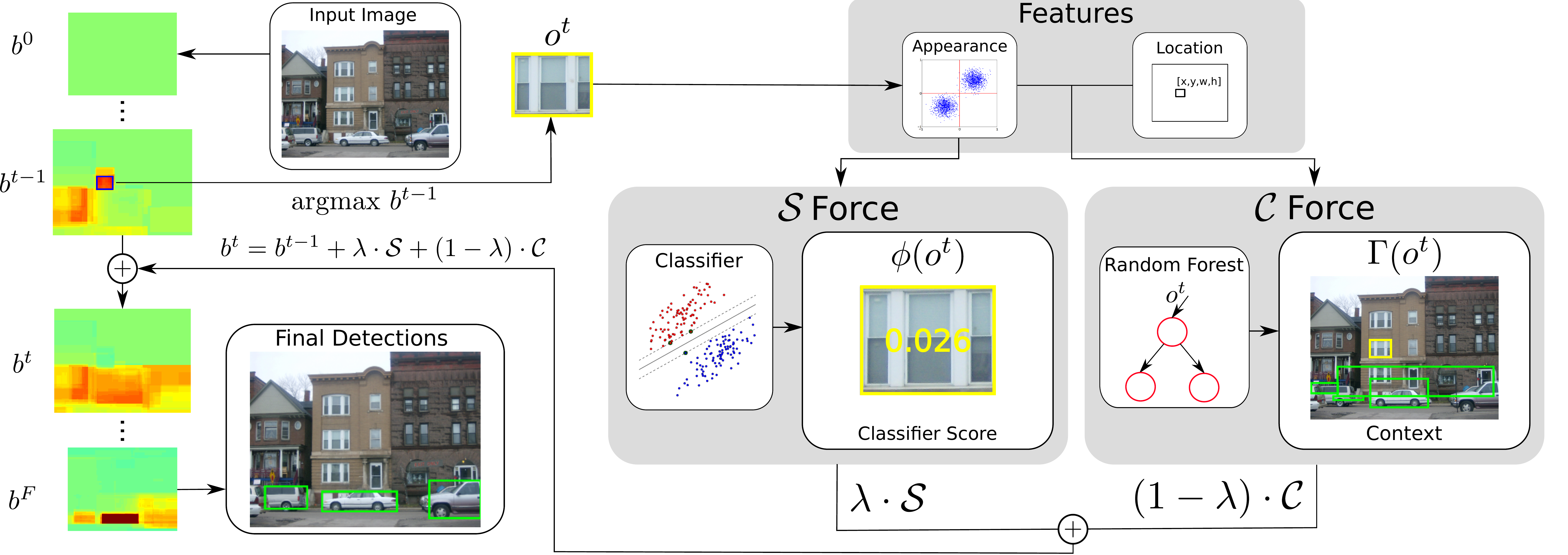}
  \vspace{-7mm}
  \end{center}
  \figshrinky
  \caption{\textbf{Search model.} \it The next observed window $o^t$ is the maximum of the current belief map $b^{t-1}$.
    The method extracts appearance and location features for $o^t$, and uses them to compute its context $\mathcal{C}$ and window classifier $\mathcal{S}$ outputs.
  Then, it combines these outputs with the current belief map $b^{t-1}$ into the next iteration's belief map $b^{t}$. 
  The final belief map $b^F$ combines all the performed observations. 
  The output detections are the observed windows with highest scores.
  \figskiny \figsmall}
  \label{fig:pipeline}
\end{figure*}

\parskiny
\paragraph{Context.}
Many works use context as an additional cue on top of object detectors, complementing the information provided by the window classifier, but without altering the search process.
Several works~\cite{felzenszwalb10pami,harzallah09iccv,murphy03nips,Torralba03,uijlings13ijcv} predict the presence of object classes based on global image descriptors, and use it to remove out-of-context false-positive detections.
The response of detectors for multiple object classes also provides context, as it enables to reason about co-occurrence~\cite{rabinovich07iccv} and spatial relations between classes~\cite{choi10cvpr, desai:iccv09,galleguillos08cvpr,heitz08eccv}.
Other works incorporate regions outside the object into the window classifier~\cite{Dalal05,li11iccv,mottaghi14cvpr,uijlings13ijcv}
%Mottaghi CVPR14 incorporates context into a part-based window classifier.
%
Divvala et al.~\cite{divvala09CVPR} analyze several context sources and their impact on object detection.

The most related work to ours is~\cite{alexe12nips}, which proposes a search strategy driven by context. Here we go beyond in several ways:
(1) They used context in an inefficient way, involving a nearest-neighbour search over all windows in all training images.
%\todo{AG: introduce NN acronym here? Used several times later.}  
This caused a large overhead that compromised the actual wall-clock speedup they made over evaluating all windows in the test image.
In contrast, we present a very efficient technique based on Random Forests, which has little overhead (sec.~\ref{sec:runtime}).
(2) While~\cite{alexe12nips} uses only context, we guide the search also by the classifier score, and learn an optimal combination of the two forces (sec.~\ref{sec:searchModel}).
(3) They perform single-view and single-instance detection, whereas we detect multiple views and multiple instances in the same image.
(4) We adopt the state-of-the-art R-CNN~\cite{girshick14cvpr} as the reference detector and compare to it, as opposed to the weaker DPM detector~\cite{felzenszwalb10pami}.
%\todo{don't we do caffe'? Clarify. AG: we do, we use Krizhevksy's network model, cite too?} 
(5) While~\cite{alexe12nips} performs experiments only on PASCAL VOC10, we also use SUN2012~\cite{xiao10cvpr}, which has more cluttered images with smaller objects.

\seckiny
\section{Search model}
\figsmall
\label{sec:searchModel}
Let $I$ be a test image represented by a set of object proposals~\cite{uijlings13ijcv}, $I=\{o_i\}^N_{i=1}$.
%Object proposals are a small set of windows, about a few thousand, likely to cover the vast majority of the objects of any class in an image~\cite{alexe12pami,uijlings13ijcv,manen13iccv}.
The goal of our method is to efficiently detect objects in $I$, by evaluating the window classifier on only a subset of the proposals.  
Our method is a class-specific iterative procedure that evaluates one window at a time.
At every iteration $t$, it selects the next window $o^{t+1}$ according to all the observations $\{o^k\}_{k=1}^{t}$ performed so far.\footnote{$o_i$ indexes through the input set of proposals $I$, whereas $o^t$ is the proposal actively chosen by our strategy in the $t$-th iteration.}
We assign a belief value $b^t(o_i, \{o^k\}_{k=1}^{t}; \Theta)$ to each object proposal $o_i$ and update it after every iteration.
This belief indicates how likely it is that $o_i$ contains the object, given all previously observed windows $\{o^k\}_{k=1}^{t}$.
Here $\Theta=\{\lambda,\sigma_\mathcal{S},\sigma_\mathcal{C}\}$ are hyperparameters and $t$ indexes the iteration.

%Our method performs sequential observations in a given image $I$ in order to guide a class-specific search strategy.  
%We act over finite window space formed by Selective Search~\cite{uijlings13ijcv} object proposals $\{o_i\}$, in the range of a few thousand. 
%The object proposals already offer good candidates likely to cover most of the objects, but most of them are redundant or uninteresting. 
%Our approach tries to efficiently find the objects by evaluating a subset of the proposals, in the range of the few hundred. 

The method starts with the belief map $b^0(o_i)=0$ $\forall o_i$, representing complete uncertainty. 
At iteration $t$, the method selects the window with the highest belief 
\begin{equation}
  \figsmall
  o^{t} = \argmax_{o_i\in I \backslash \{o^k\}_{k=1}^{t-1}} b^{t-1}(o_i, \{o^k\}_{k=1}^{t-1}; \Theta)
\end{equation}
We avoid repetition by imposing $o^{t} \neq o^{k}$, $\forall k < t$.
The starting window $o^1$ is the average of all the ground-truth bounding-boxes in the training set. 

%the method creates an individual belief map $\tilde{b}^t(o_i; \Theta)$ for the current iteration. 
At each iteration $t$, the method obtains information from the new observation $o^t$ and it updates the belief values of all windows as follows
\figsmall
\begin{multline}
  b^{t}(o_i, \{o^k\}_{k=1}^{t};\Theta) = b^{t-1}(o_i, \{o^k\}_{k=1}^{t-1};\Theta)\\ + \lambda\cdot \mathcal{S}(o_i,o^t; \sigma_\mathcal{S}) + (1-\lambda)\cdot \mathcal{C}(o_i,o^t; \sigma_\mathcal{C})
\end{multline}

The observation $o^t$ provides two kind of information: the context $\mathcal{C}$ and the classifier score $\mathcal{S}$ (explained below).
These are linearly combined with a mixing parameter $\lambda\in[0,1]$.
Fig.~\ref{fig:pipeline} illustrates our pipeline.

\parskiny
\paragraph{Context force $\mathcal{C}$}
points to areas of the image likely to contain the object, relative to the observation $o^t$.
It uses the statistical relation between the appearance and location of training windows and their position relative to the objects.
The context force may point to any area of the image, even those distant from $o^t$.
For the car detection example, if $o^t$ contains a patch of building, $\mathcal{C}$ will point to windows far below it, as cars are below buildings (fig.~\ref{fig:searchEx},~\ref{fig:rfcnn}).
If $o^t$ contains a patch of road, $\mathcal{C}$ will propose instead windows next to $o^t$, as cars tend to be on roads (fig.~\ref{fig:searchEx}). 
%\todo{what about using fig 4 and 6? AG: I used fig. 6, as it shows the building example. I wouldn't use fig.4, as it contains explicitly displacement vectors and it might be confussing now.}

The heart of the force $\mathcal{C}$ is a \emph{context extractor} $\Gamma$.
Given the appearance and location of $o^t$, $\Gamma$ returns a set of windows $\Gamma(o^t)=\{w_j\}_{j=1}^J$ (not necessarily object proposals).
These windows cover locations likely to contain objects of the class, as learned from a set of training windows and their relative position to the objects in their own images. We explain how our context extractor works in sec.~\ref{sec:contextExtractor}.

We can now define $\mathcal{C}$ as 
\figsmall
\begin{equation}
\figsmall
  \mathcal{C}(o_i,o^t;\sigma_\mathcal{C}) = \sum_{w_j\in \Gamma(o^t)} K(w_j, o_i; \sigma_\mathcal{C})
\end{equation}
It gives high values to object proposals close to windows in $\Gamma(o^t)$, as we expect these windows to be near objects. 
The influence of the windows in $\Gamma(o^t)$ is weighted by a smoothing kernel
\figsmall
\begin{equation}
\figsmall
  \label{eq:kernel}
  K(w,o;\sigma) = e^{-(1-\text{IoU}(w,o))^2/(2\sigma^2)}
\end{equation}
This choice of kernel assumes smoothness in the presence of an object for nearby windows.
Indeed, adjacent windows to a window containing an object will also contain part of the object.  
The further apart the windows are, the lower the probability of containing the object is.
We use the inverse overlap 1 - intersection-over-union~\cite{Everingham10} (IoU) as distance between two windows.

\begin{figure}
  \begin{center}
  \includegraphics[width=0.45\textwidth]{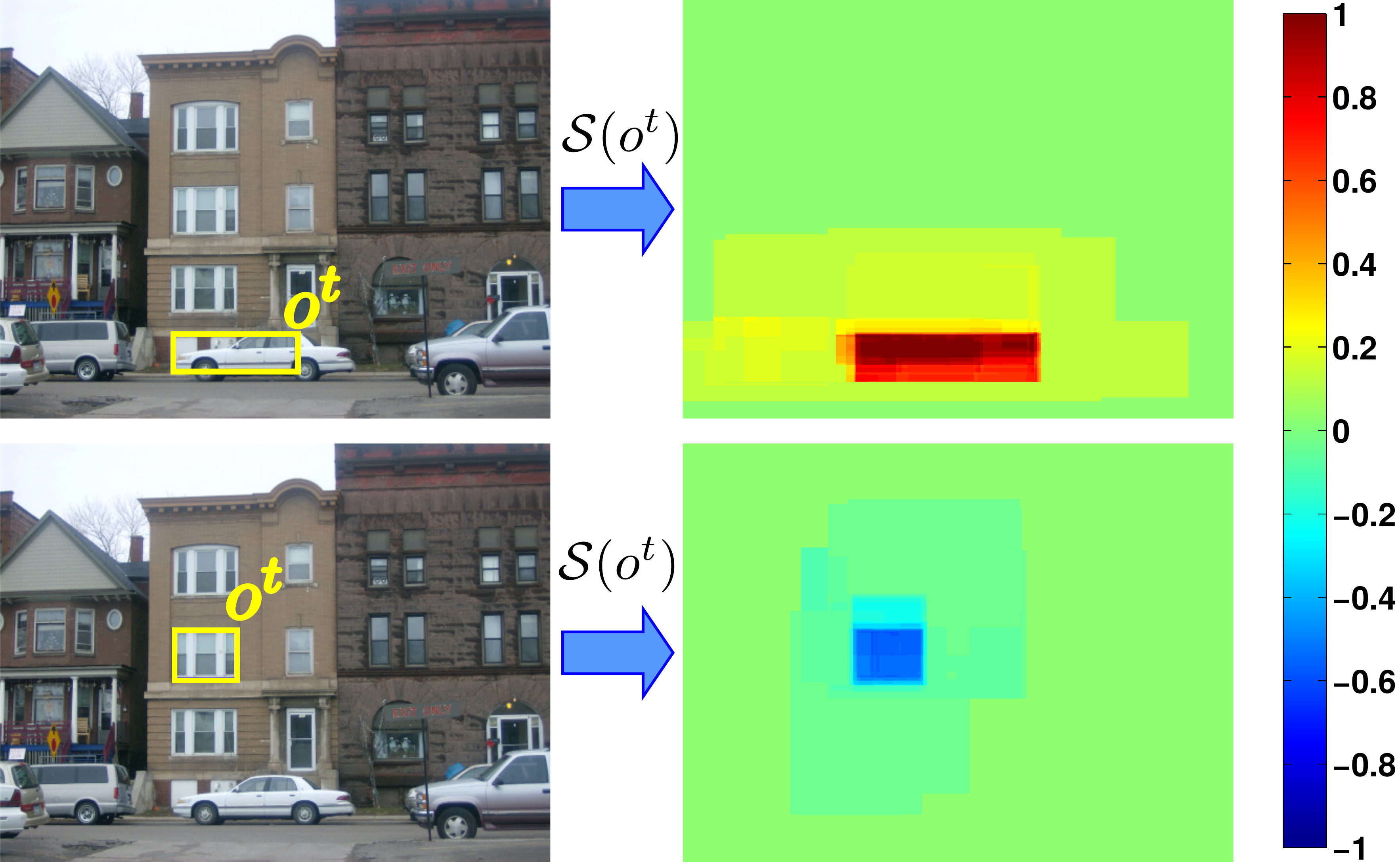}
  \end{center}
  \figskiny
  \caption{\small \textbf{Classifier score force $\mathcal{S}$.} \it (Left) Image and observation $o^t$. 
  (Right) Belief map produced by $\mathcal{S}$. Colours correspond to belief values. \figskiny}  
  \label{fig:ScoreForce}
\end{figure}

\parskiny
\paragraph{Classifier score force $\mathcal{S}$}
attracts the search to the area surrounding the observation $o^t$ if it has high classifier score, while pushing away from it if it has low score:
\begin{equation}
  \mathcal{S}(o_i,o^t;\sigma_\mathcal{S}) = K(o_i,o^t;\sigma_\mathcal{S}) \cdot (\phi(o^t)-0.5)
\end{equation}
where $\phi(o^t) \in [0,1]$ is the window classifier score of $o^t$
(sec.\ref{sec:classifierScores} details our choice of window classifier).
We translate $\phi(o^t)$ into the $[-0.5,0.5]$ range and weight it using the smoothing kernel~(\ref{eq:kernel}). 
Therefore, $\mathcal{S}$ operates in the surroundings of $o^t$, spreading the classifier score to windows near $o^t$.
When $\mathcal{S}$ is positive, it attracts the search to the region surrounding $o^t$.
For example, if $o^t$ contains part of a car, it will probably have a high classifier score (fig.~\ref{fig:ScoreForce}).
Then $\mathcal{S}$ will guide the search to stay in this area, as some nearby window is likely to contain the whole car.
On the other hand, when $\mathcal{S}$ values are negative, it has a repulsive effect.
It pushes the search away from uninteresting regions with low classifier score, such as background (sky, buildings, etc).

\begin{figure}
  \begin{center}
  \hspace{-1mm}
  \begin{subfigure}[b]{0.18\textwidth}
    \includegraphics[width=\textwidth]{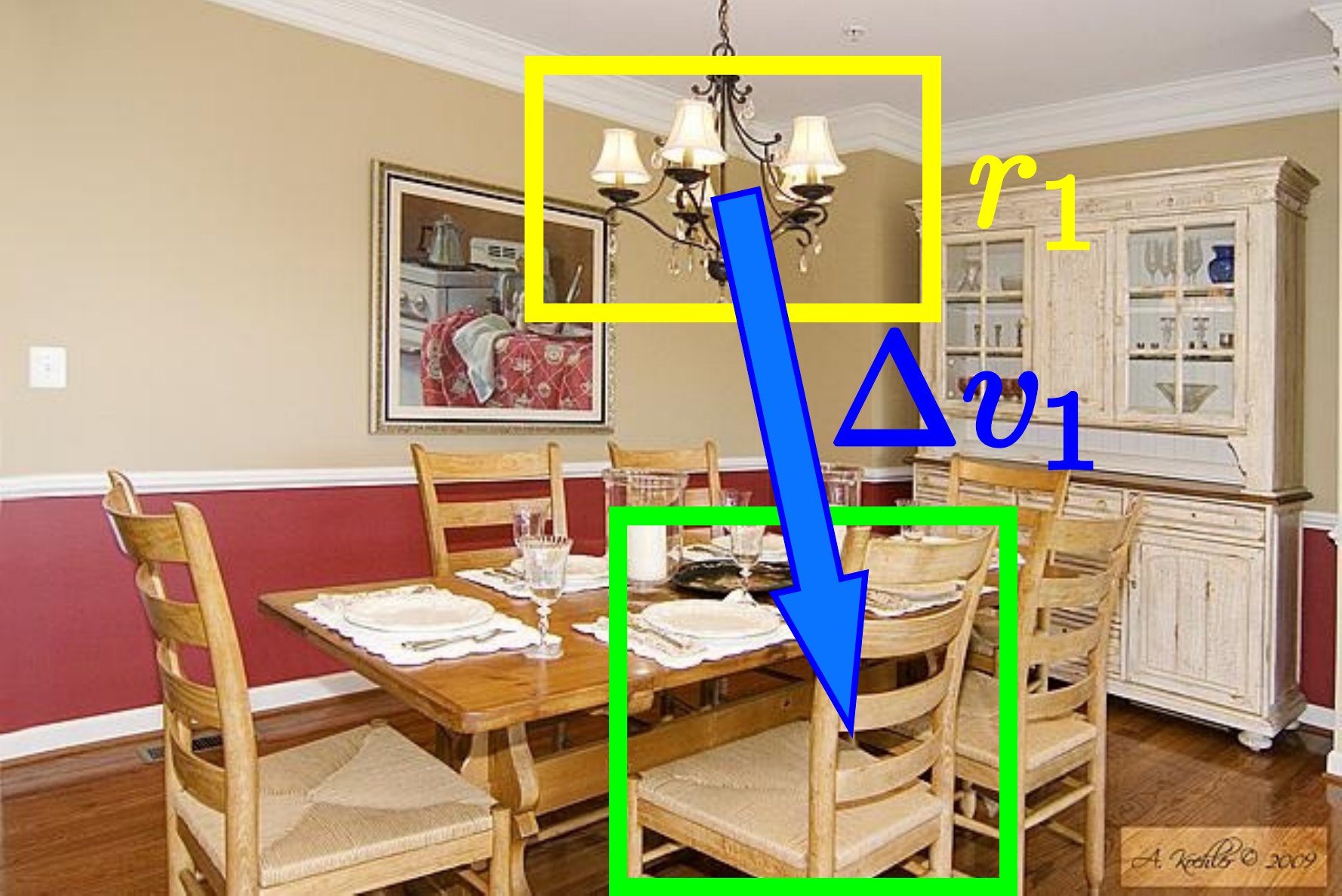}
    \caption{}
    \label{fig:contextExtractora}
  \end{subfigure}
  ~\hspace{-1mm} 
  %\vspace{0.5mm}
 \begin{subfigure}[b]{0.18\textwidth}
    \includegraphics[width=\textwidth]{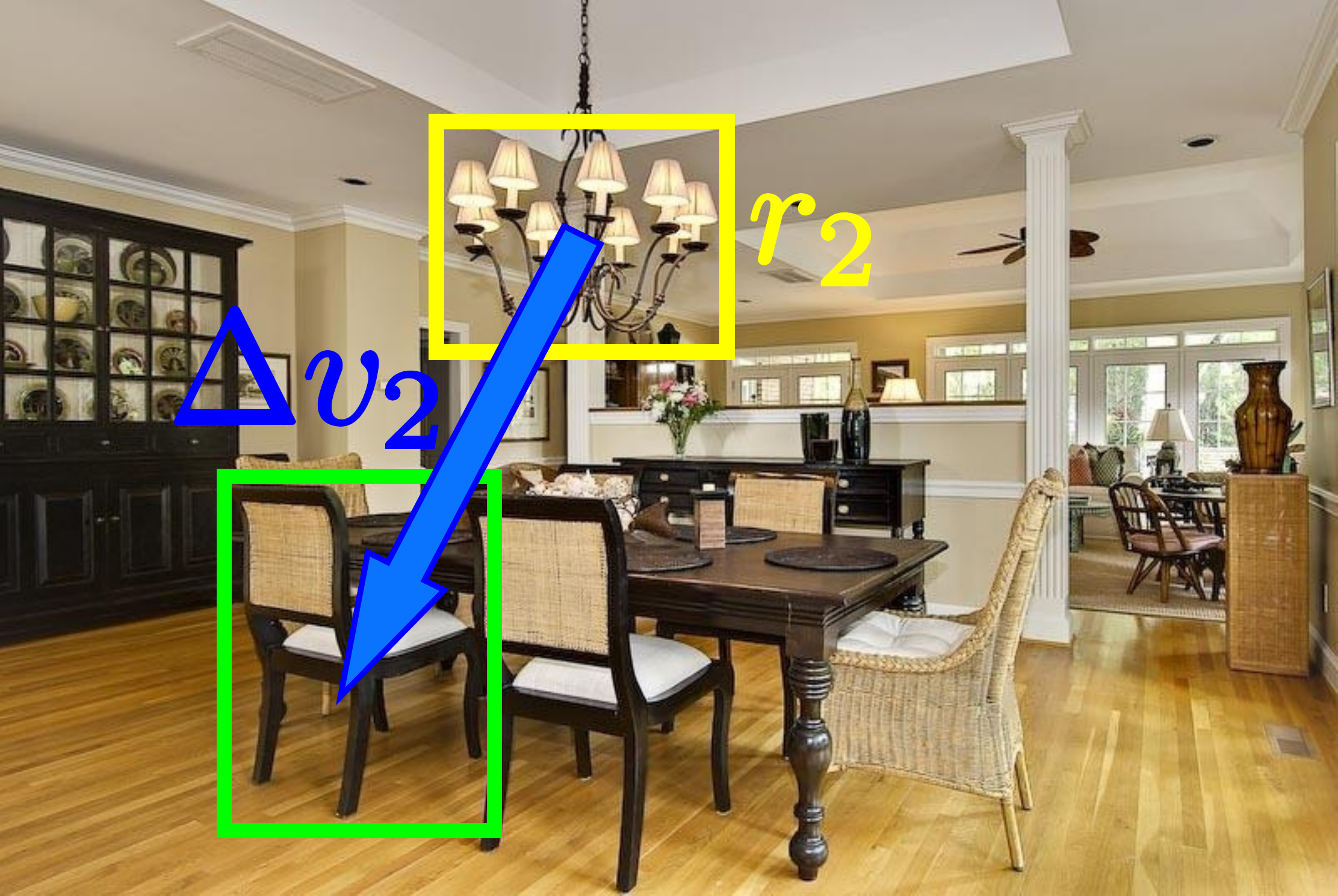}
    \caption{}
    \label{fig:contextExtractorb}
  \end{subfigure}
  \begin{subfigure}[b]{0.36\textwidth}
    \includegraphics[width=\textwidth]{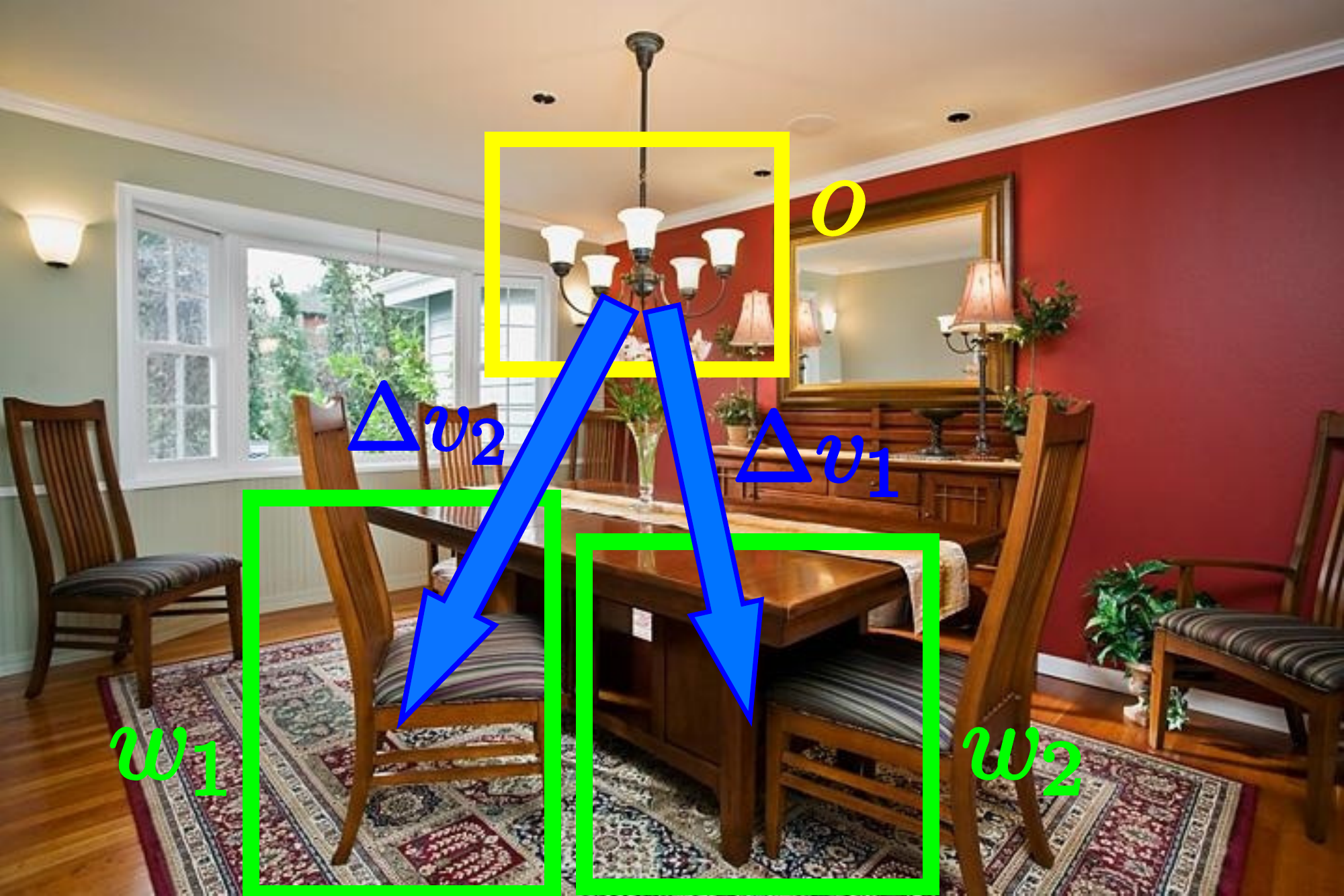}
    \caption{}
  \end{subfigure}
  \end{center}
  \figskiny
  \figshrinky
  \caption{\textbf{Context extractor.} \it (a,b) The displacement vectors $\Delta v_1$ and $\Delta v_2$ of training samples $r_1$ and $r_2$ point to their respective ground-truth objects. 
  (c) By applying $\Delta v_1$ and $\Delta v_2$ to test observation $o$, we obtain displaced windows $w_1$ and $w_2$, covering likely locations for the object with respect to $o$. \figskiny \figsmall}
  \label{fig:contextExtractor}
\end{figure}
 
%When $o^t$ is a highly scored window (containing a car), $\mathcal{S}$ gives high belief values to surrounding windows, creating an attracting region.
%The search will prioritize visiting this region, as it is likely that it contains a car.
%If $o^t$'s score is low instead, $\mathcal{S}$ creates a repulsing region, giving low belief values to nearby windows.
%Hence, the search will probably not visit these windows, as they are unlikely to contain a car.

%The force $\mathcal{S}$ is based on the score of the classifier, and it can act in two ways.  
%For a highly scored window, similar windows in its neighborhood should have high belief values. 
%They are likely to have a high classifier score, due to the proximity to a high score window and the smoothness of the classifier. 
%Analogously, the  belief value should be low near windows with low scores. 
%The model allows for negative beliefs in order to avoid wasting computation time in uninteresting areas, highly unlikely to contain objects. 
%\paragraph{Smoothing kernel $K$:}
%i.e., the Gaussian kernel on the distance, where $\sigma$ is the standard deviation. This kernel is commonly used in Kernel Density Estimation~\cite{hwang94tsp} and Mean Shift~\cite{Comaniciu}.
%In our case, however, the Euclidean distance in the 4D space of windows is not appropiate, as it does not express well the actual difference between windows.
%This acts as a finite suport approximation of the distance between two windows. 

\parskiny
\section{Context extractor}
\label{sec:contextExtractor}
\figsmall
Given an input observation $o$ in the test image, the context extractor $\Gamma$ returns a set of windows $\Gamma(o)=\{w_j\}_{j=1}^J$ covering locations likely to contain objects.

%exploiting the statistical relation between the appearance and location of training windows and their position relative to the objects.
%The output of $\Gamma(o^t)$ is a distribution of windows representing where the object should be, given that we have observed window $o^t$.
The context extractor is trained from the same data as the window classifier, i.e. images with annotated object bounding-boxes. 
Hence our approach requires the same annotation as standard object detectors~\cite{cinbis13iccv,Dalal05:thomas,felzenszwalb10pami,girshick14cvpr, harzallah09iccv, MalisiewiczICCV11,  uijlings13ijcv,wang13iccv}.
The training set consists of pairs $\{(r_n,\Delta v_n)\}_{n=1}^N$; $r_n$ is a proposal from a training image, and $\Delta v_n$ is a 4D displacement vector, which transforms $r_n$ into the closest object bounding-box in its training image (fig.~\ref{fig:contextExtractor}a-b). 
Here index $n$ runs over all object proposals in all the training images.
For 500 images and 3200 proposals per image, $N = 500\cdot 3200 = 1'600'000$.

Given the observation, the context extractor regresses a displacement vector $\Delta v$ pointing to the object
%\footnote{
%Note the difference with Hough Forests~\cite{gall09cvpr}, which uses random forests within the object class model: the leaves form an implicit codebook optimized for hough-style object detection. Instead we use random forests to model context in order to guide the search.}.
For robustness, the context extractor actually outputs a set of displacement vectors $\{\Delta v_j\}_{j=1}^J$, to allow for some uncertainty regarding the object's location. 
Then it applies $\{\Delta v_j\}_{j=1}^J$ to $o$, obtaining a set of displaced windows on the test image: $\Gamma(o) = \{o + \Delta v_j\}_{j=1}^J$.
The windows in $\Gamma(o)$ indicate expected locations for the object in the test image.
Note that they may be any window, not necessarily object proposals. 

\parshrinky
\paragraph{Random Forests.}
%\todo{AV: section changed a lot, some todos where too outdated to keep}
We use Random Forests (RF)~\cite{breiman01ml} as our context extractor. 
A RF is an ensemble of $J$ binary decision trees, each tree inputs the window $o$ and outputs a displacement vector $\Delta v_j$. The final output of RF are all displacement vectors $\{\Delta v_j\}_{j=1}^J$ produced by each tree.
% $\Gamma$, regressing $\{\Delta v_j\}_{j=1}^J$ given test window $o$.
%Given test window $o$, we regress $\{\Delta v_l\}_{l=1}^L$ using RF as our regressor. 
%We regress $\{\Delta v_l\}_{l=1}^L$ given a particular test window $o$. 
%A RF is an ensemble of $J$ binary decision trees, whose individual outputs are combined to produce the final output. 
%In our case, each tree inputs window $o$ and outputs a displacement vector $\Delta v$.
%The overall output of the RF is the set of displacement vectors $\{\Delta v_j\}_{j=1}^J$ output by each tree. 
%\todo{overload of delta v, now subscript points to tree ... bad! AG: I think it is an acceptable overload. The context extractor returns $J$ displacement vectors, each tree regresses one and we have $J$ trees.}

\begin{figure}[t]
  \begin{center}
  \includegraphics[width=0.7\columnwidth]{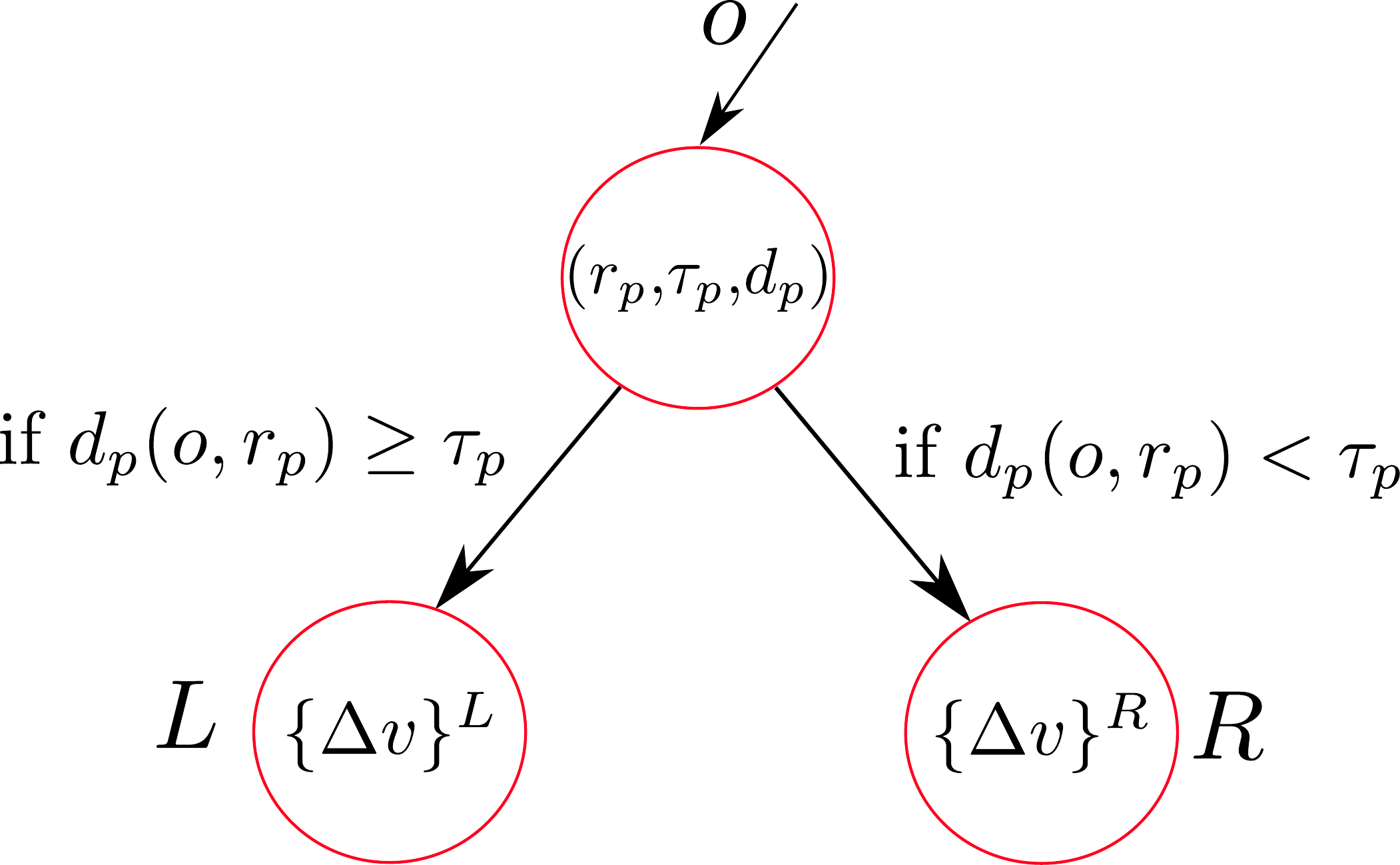}
  \end{center}
  \figskiny
  \figsmall
  \caption{\textbf{Internal node at test time.} \it 
  The test compares distance $d_{p}(o,r_p)$ between test window $o$ and training sample $r_p$ with the threshold $\tau_p$. \figskiny \figsmall} 
  
   % A test sample $o$ arrives and the distance $d_i$ with the node sample $o^{tr}_n$ is computed. The direction of the propagation depends on the threshold $\tau$. \figskiny}
  \label{fig:RFTest}
\end{figure}

RF have been successfully applied in several learning problems such as classification, regression or density estimation~\cite{criminisi2011}. 
RF in computer vision~\cite{criminisi2011,fanelli11cvpr,gall09cvpr,girshick11iccv,montillo09icip} typically use axis-aligned separators (thresholding on one feature dimension) as tests in the internal nodes.
However, we found that tests on distances to training samples perform better in our case, as they are more informative.
%Since we encode our appearance features in a binary space (sec.~\ref{sec:implementationDetails}), some features are single bits, which are not discriminative enough. 
%\todo{we'll see if we report it. AG: we don't.}
%\todo{WHY DOES IT DO THAT? Because each test is much more informative in that case. AG: explained.}
Hence, we build our RF based on distance tests.
This is related to Proximity Forests~\cite{ohara13wacv}, although~\cite{ohara13wacv} uses RF for clustering, not geometric regression.

When the test window $o$ goes down a tree, it traverses it from the root to a leaf guided by tests in the internal nodes.
At each node $p$ the decision whether $o$ goes left or right is taken by comparing the distance $d_{p}(o,r_p)$ between $o$ and a pivot training point $r_p$ to the threshold $\tau_p$. 
The test window $o$ proceeds to a left child if $d_{p}(o,r_p)\geq \tau_p$ or to the right child otherwise (fig.~\ref{fig:RFTest}).
This process is repeated until a leaf is reached.
Each leaf stores a displacement vector, which the tree returns. 
%Finally, the output of the tree is the mediod displacement vector of the reached leaf, i.e. the displacement vector closest to the mean of all displacement vectors contained in the leaf.
The triplet $(r_p$,$\tau_p$,$d_{p})$ at each internal node is chosen during training
(the process can choose between two different distance functions, sec.~\ref{sec:implementationDetails}).

\parskiny
\parskiny
\paragraph{RF training.}
For each class, we train one RF with $J=10$ trees.
%We train $J=10$ trees per RF, and one RF per class. 
To keep the trees diverse, we train each one on windows coming from a random sub-sample of 40 training images. 
%\todo{AG: Vitto's note that I don't understand.}
As shown in~\cite{criminisi2011}, this procedure improves generalization.  
%\todo{AV: What we do is not bagging} 
We construct each tree by recursively splitting the training set at each node. We want to learn tests in the internal nodes such that leaves contain samples with a compact set of displacement vectors. 
This way a tree learns to group windows using features that are predictive of their relative location to the object.
%We train each tree so the set of displacement vectors in each leaf is compact.
%Each training sample $(o^{tr}_n,\Delta v_n)$ is composed of a training window $o^{tr}_n$ and its associated displacement vector $\Delta v_n$.
%The injection of randomness in the training data selection (bagging) has been shown to improve the accuracy of regression with RF~\cite{breiman01ml,criminisi2011}.
%We randomly select a subset of training samples $\{(r_k,\Delta v_k)\}_{k=1}^K$ to learn each tree, using all available proposals for 40 different images ($K<N$).
To create an internal node $p$, we need to select a triplet $(r_p$,$\tau_p$,$d_{p})$, which defines our test function. 
Following the extremely randomized forest framework~\cite{Moosman06} we generate a random set of possible triplets.
We then pick the triplet that achieves maximum information gain:
\figsmall
%\todo{this text is very terrible. $d_i$ was not even in the combination list 3 lines ago!!}
\begin{equation}
  \figsmall
IG = H(S) - \dfrac{|S^L|}{|S|}\cdot H(S^L) - \dfrac{|S^R|}{|S|}\cdot H(S^R),
\end{equation}
where $S$ is the set of training samples at the node and $L,R$ denote the left and right children with samples $S^L$ and $S^R$, respectively. 
$H$ is the approximated Shannon entropy of the 4D displacement vectors in $S$:
we first compute a separate histogram per dimension, and then sum their individual entropies~\cite{hall93aism}. 
Finally, we keep in each leaf the mediod displacement vector of the training samples that fell into it.
%\todo{explain why this bagging, ref. to some tutorial or textbook showing it's beneficial to performance due to more rnd and so on. AG: we don't do bagging, but extremely randomized forest framework.}

\label{sec:classifierScores}
\begin{figure}[t]
\begin{center}
    \includegraphics[width=\columnwidth]{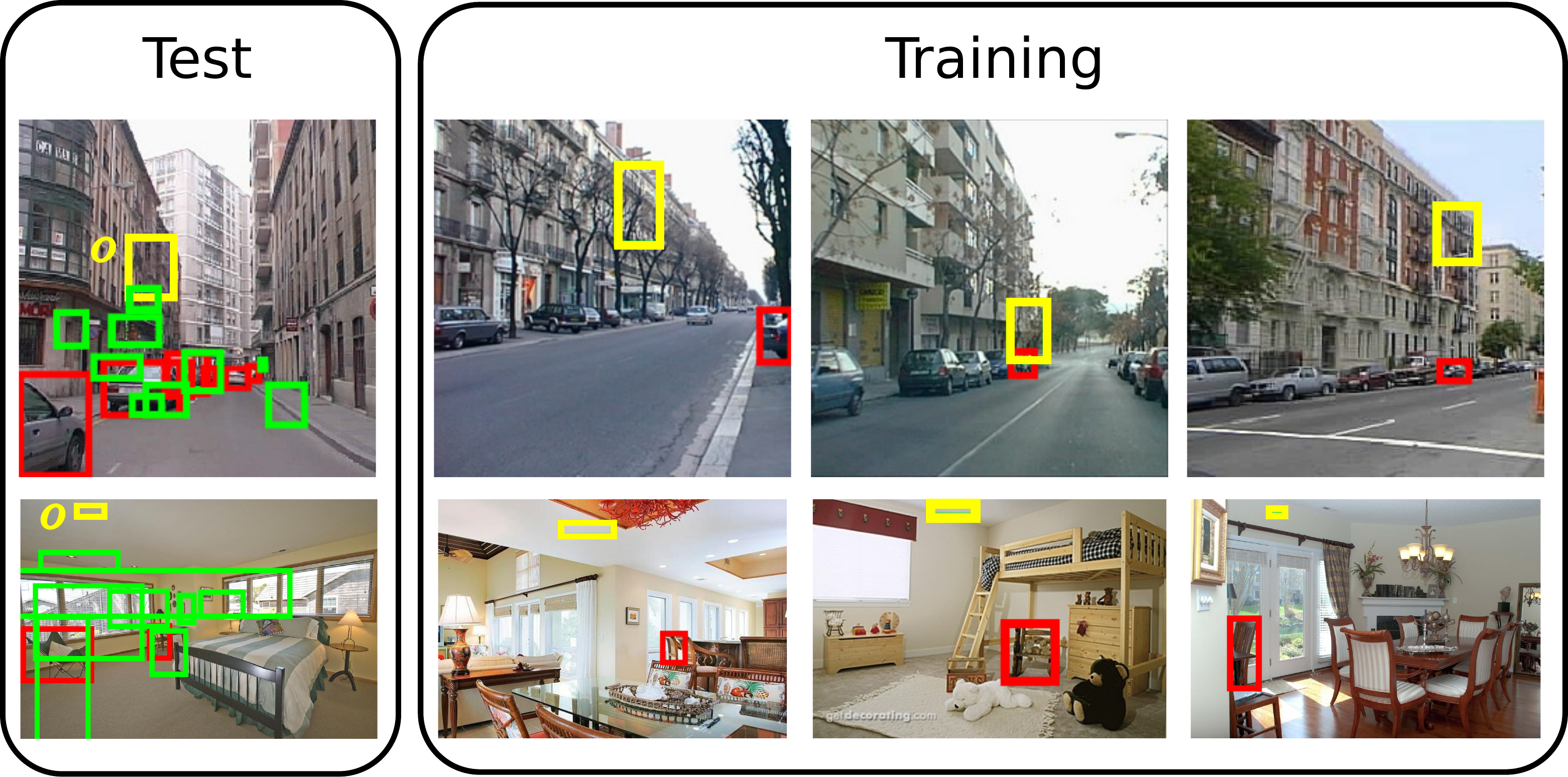}
\end{center}
\figshrinky
\caption{\small \textbf{RF examples.} \it  (Test) Example image and observation $o$ for classes car (top) and chair (bottom). 
  Windows displaced by the displacement vectors regressed by
  the RF are in green, whereas ground-truth objects are in red.
  (Training) Training samples (yellow) in leaves reached by $o$ when input into our RF and associated
ground-truth objects (red). \figskiny } 
\label{fig:rfcnn}
\end{figure}

Fig.~\ref{fig:rfcnn} shows examples test windows passed through the RF, along with example training windows in leaves they reach.
Note how these training windows are similar in appearance to the test window, and produce displaced windows  covering areas likely to contain objects of that class in the test image. 
This demonstrates that RF is capable of learning the relation between the appearance of a window and its position relative to the object.
%\todo{terrible: does it look like all training samples? Or like the 'retrieved' ones? AG:explained.}
%Moreover, the resulting displaced windows cover areas where instances of the classes are likely to be.

\parskiny
\parskiny
\paragraph{Efficiency.}
A key benefit of RF over a simple nearest-neighbour regressor~\cite{alexe12nips} is computational efficiency. 
Since the observation $o$ is only compared to at most as many pivot points as the depth of the tree, the runtime of RF grows only logarithmically with the size of the training set. 
In contrast, the runtime of nearest-neighbour grows linearly, since it compares $o$ to all training samples.  This makes a substantial different in runtime in practice (sec.~\ref{sec:runtime}).
%are efficient regressors due to their logarithmic complexity at test time.
%An alternative context extractor using a nearest-neighbour search over a database of training images as in~\cite{alexe12nips} has linear complexity instead.  
%This leads to a substantial overhead, whereas in our case the overhead is small.
%\todo{somewhere in this section you need to give SOUL: reasons why this RF makes sense: why should it retrieve windows that look similar to training ones? Also, why fast? Give some sense of core complexity vs NN}

\seckiny
\section{Classifier scores}
\figsmall
As our method supports any window classifier $\phi$, we demonstrate it
on two very different ones: R-CNN~\cite{girshick14cvpr} and UvA~\cite{uijlings13ijcv}.

\parskiny
\parskiny
\paragraph{R-CNN.}
This detector is based on the CNN model of~\cite{krizhevsky12nips}, which achieved winning results on the ILSVRC-2012 image classification competition~\cite{ilsvrc12}. The CNN is then fine-tuned from a image classifier to a window classifier on ground-truth bounding-boxes. Finally, a linear SVM classifier is trained on normalized 4096-D features, obtained from the 7th layer of the CNN. 
We use the R-CNN implementation provided by the authors~\cite{girshick14cvpr}, based on \textit{Caffe}~\cite{jia13caffe}.

\parskiny
\parskiny
\paragraph{UvA.}
The Bag-of-Words technique of~\cite{uijlings13ijcv} was among the best detectors before CNNs.
A window is described by a 3x3 spatial pyramid of bag-of-words.
The codebook has 4096 words and is created using Random Forest on PCA-reduced dense RGB-SIFT~\cite{sande10pami} descriptors. Overall, the window descriptor has 36864 dimensions.
The window classifier is an SVM with a histogram intersection kernel on these features.
We use the implementation provided to us kindly by the authors~\cite{uijlings13ijcv}.

For both methods, we fit a sigmoid to the outputs of the SVM to make the classifier score $\phi$ lie in $[0,1]$.

\seckiny
\section{Technical details}
\label{sec:implementationDetails}

\parskiny
\paragraph{Object proposals.}
We use the fast mode of Selective Search~\cite{uijlings13ijcv}, giving 3200 object proposals per image on average. These form the set of windows $o_i$ visible to our method (both to the context extractor and window classifier).
Note how both the R-CNN and UvA detectors as originally proposed~\cite{girshick14cvpr,uijlings13ijcv} also evaluate their window classifier on these proposals.
%
%We take at most 5000 proposals per image, which results in an average of 3224 proposals per image.
%This set of object proposals achieve a mean Average Best Overlap (MABO)~\cite{uijlings13ijcv} of 0.714 in the SUN2012~\cite{xiao10cvpr} dataset, which is comparable to the MABO reported in~\cite{uijlings13ijcv} for PASCAL VOC 2012~\cite{Everingham10}.

\parshrinky
\paragraph{Features and distances for context extractor.}
We represent a proposal by two features: location and appearance.
A proposal location $[x/W,y/H,w/W,h/H]$ is normalized by image width $W$ and height $H$.
Here $x,y,w,h$ are the top-left coordinates, width and height of the proposal.
The distance function is the inverse overlap $1-\text{IoU}$.

The appearance features used by the context extractor match those in the window classifier.
We embed the 4096-dimensional CNN appearance features~\cite{jia13caffe} in a Hamming space with 512 bits using~\cite{GongCVPR11}.
This reduces the memory footprint by 256$\times$ (from 131072 to just 512 bits per window).
It also speeds up distance computations, as the Hamming distance between these binary strings is $170\times$ faster than L2-distance in the original space.
We do the same for the Bag-of-Words features of~\cite{uijlings13ijcv}.
These Hamming embeddings are used only for the context extractor (sec.~\ref{sec:contextExtractor}). The window classifiers work on the original features (sec.~\ref{sec:classifierScores}).

\begin{figure*}
  \begin{center}
  \begin{subfigure}[b]{0.3\textwidth}
    \includegraphics[width=\textwidth]{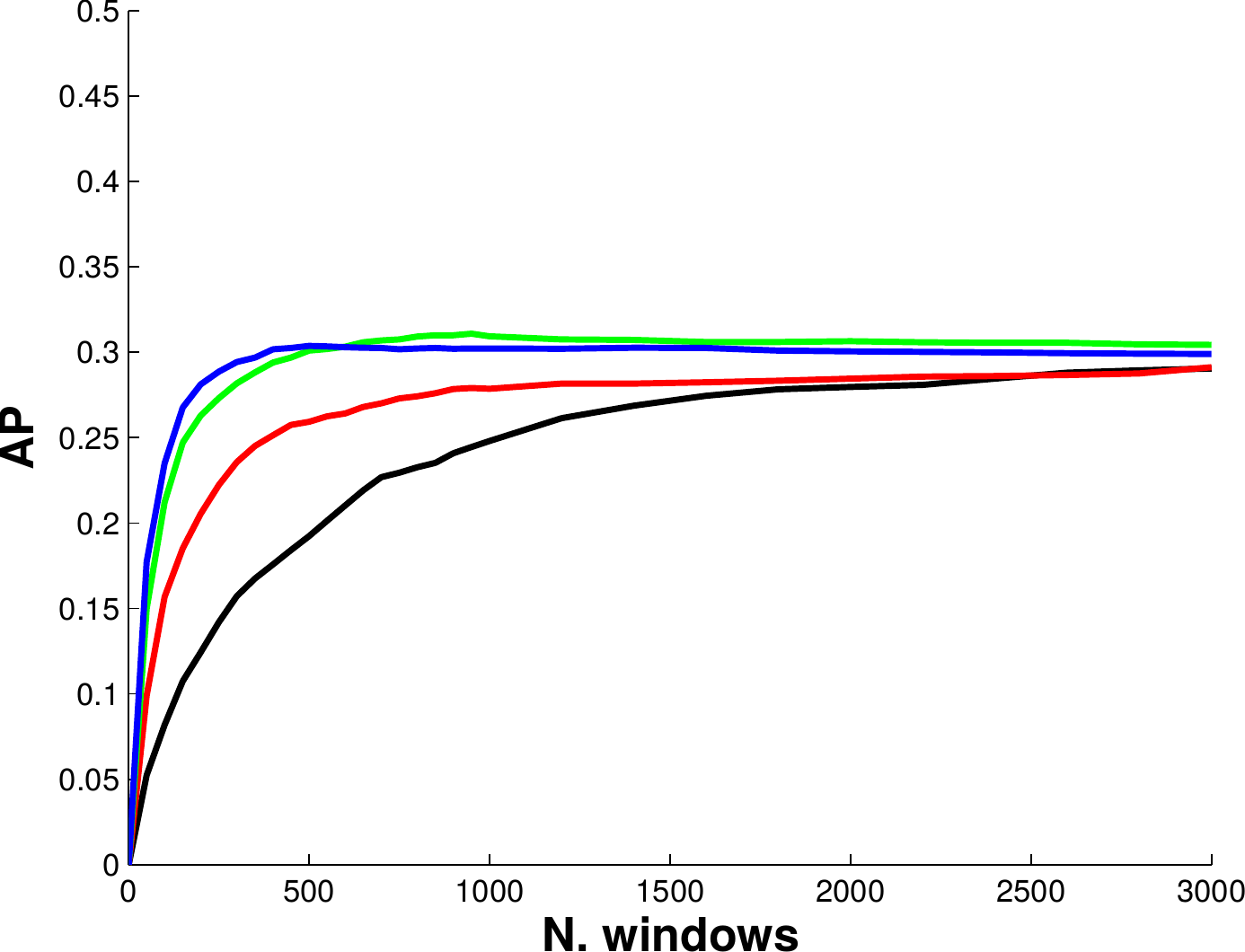}
    \caption{\small Chair}
  \end{subfigure}
  ~
 \begin{subfigure}[b]{0.3\textwidth}
    \includegraphics[width=\textwidth]{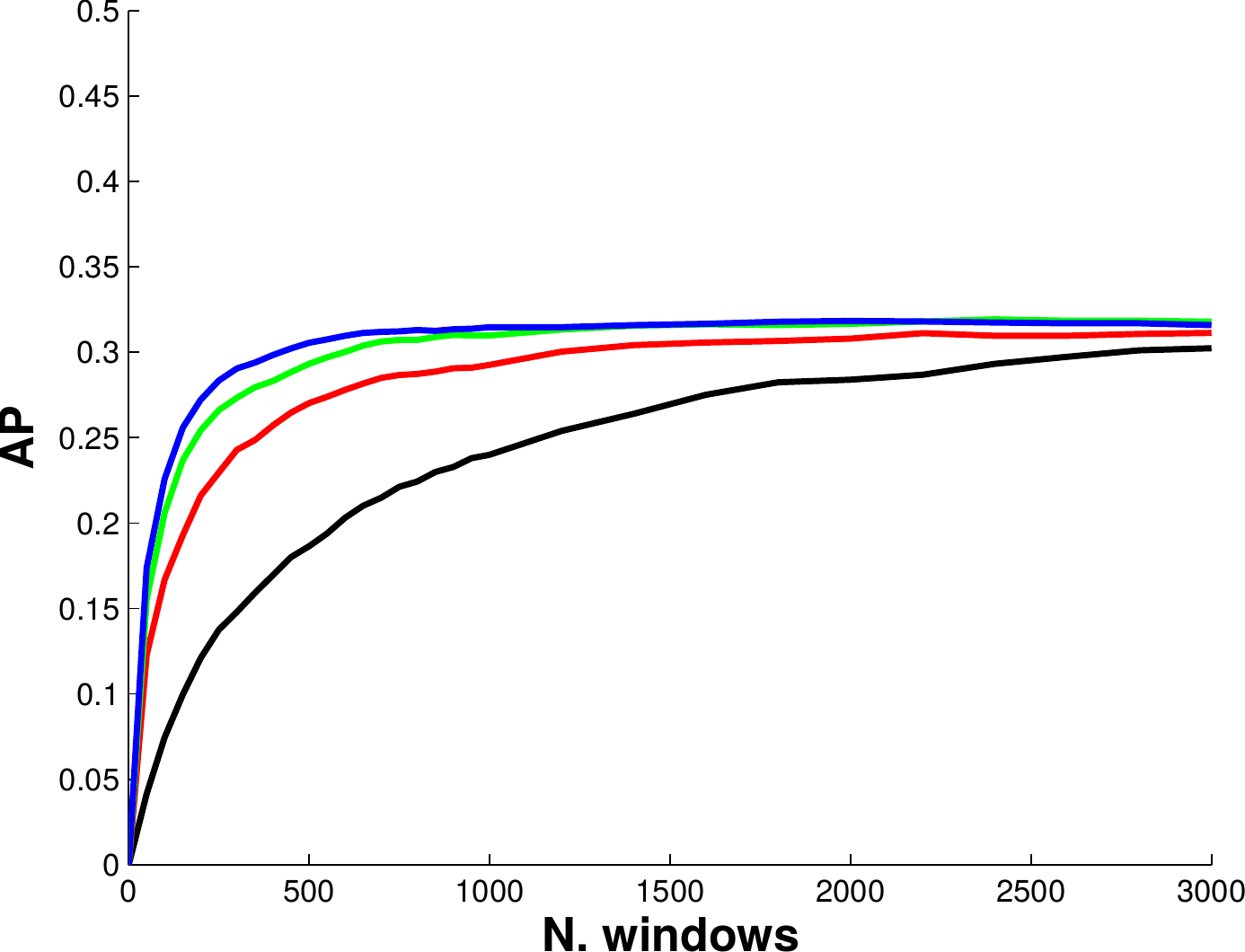}
    \caption{\small Person}
  \end{subfigure}
  ~
  \begin{subfigure}[b]{0.3\textwidth}
    \includegraphics[width=\textwidth]{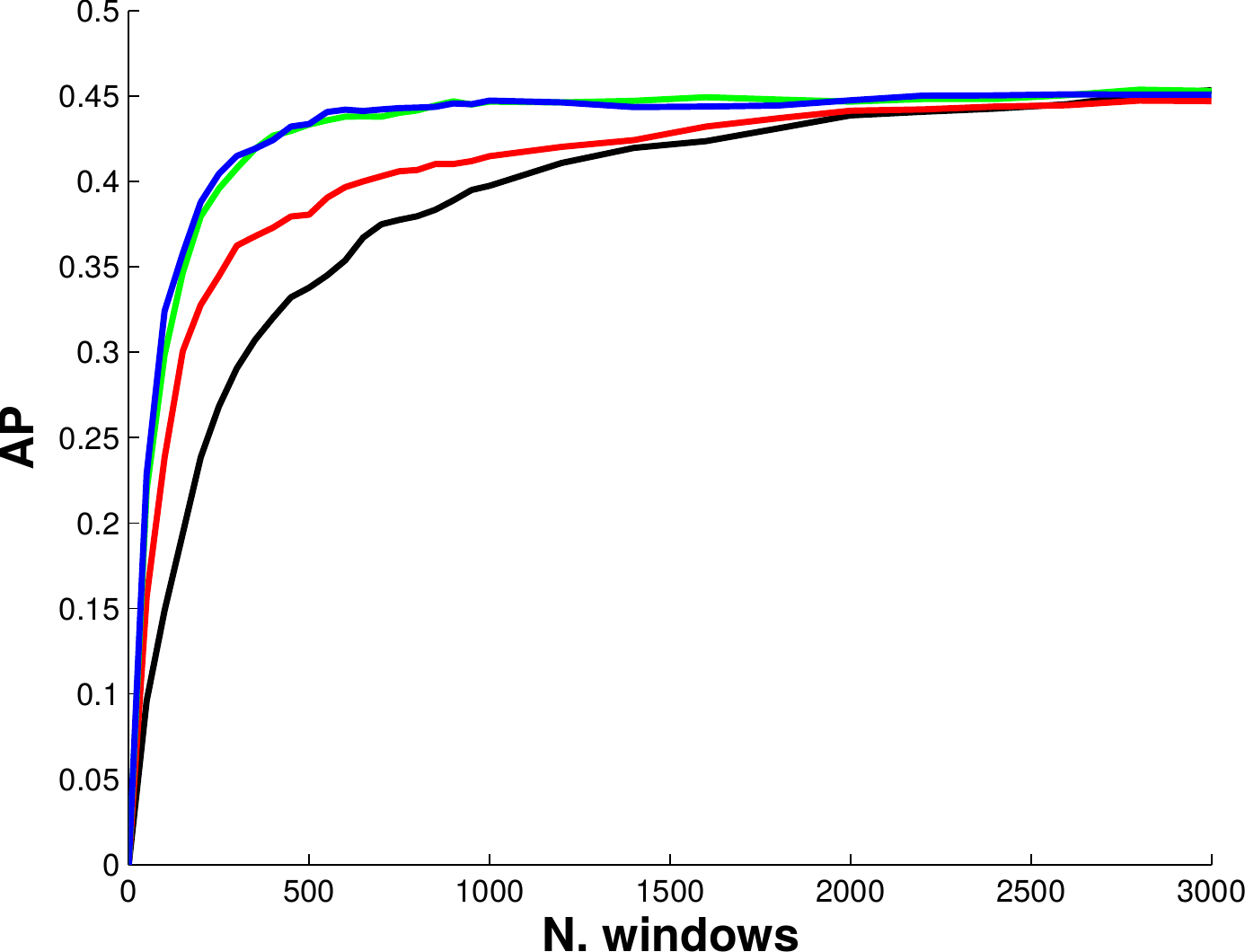}
    \caption{Car}
  \end{subfigure}
~
  \begin{subfigure}[b]{0.3\textwidth}
    \includegraphics[width=\textwidth]{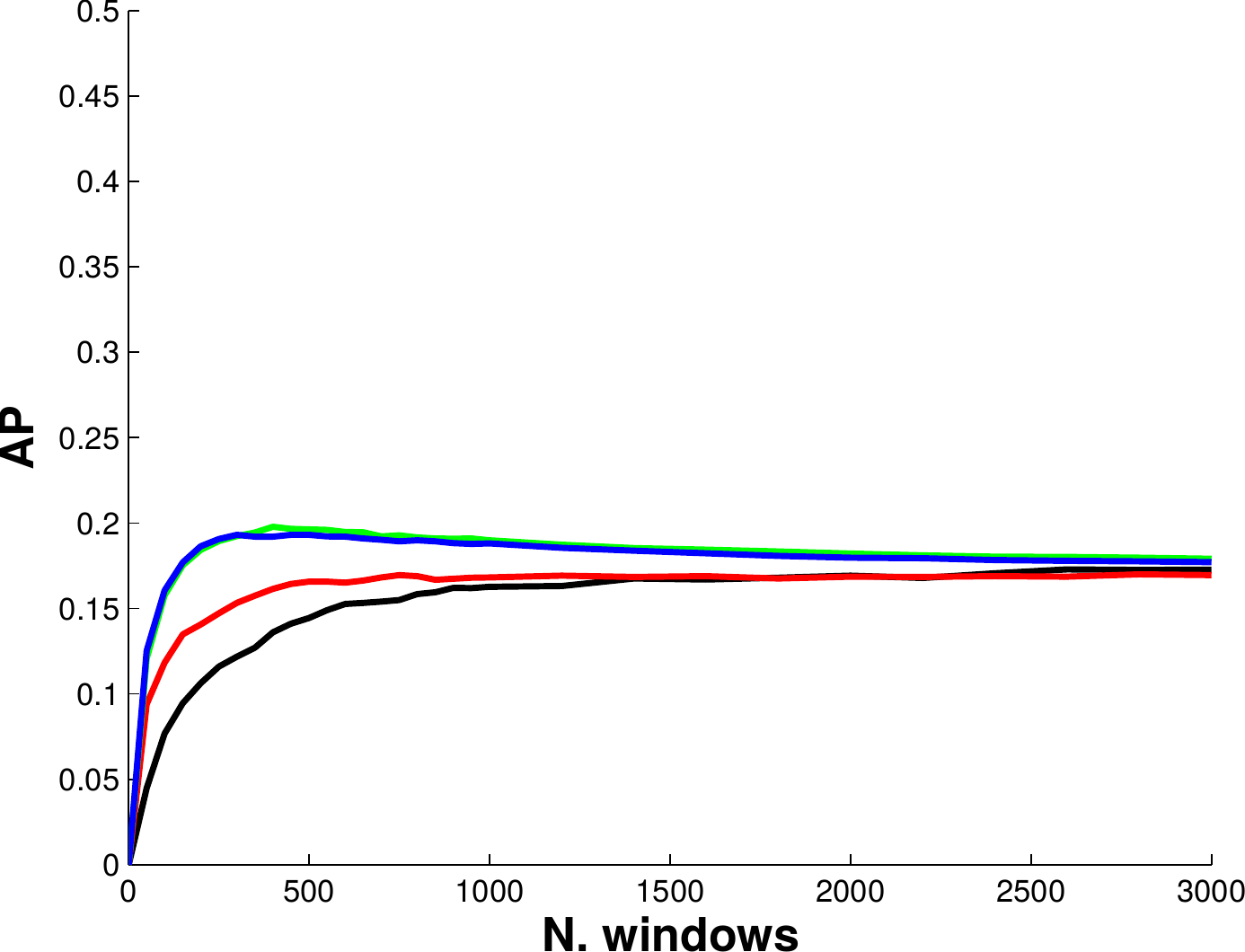}
    \caption{Door}
  \end{subfigure}
~
  \begin{subfigure}[b]{0.3\textwidth}
    \includegraphics[width=\textwidth]{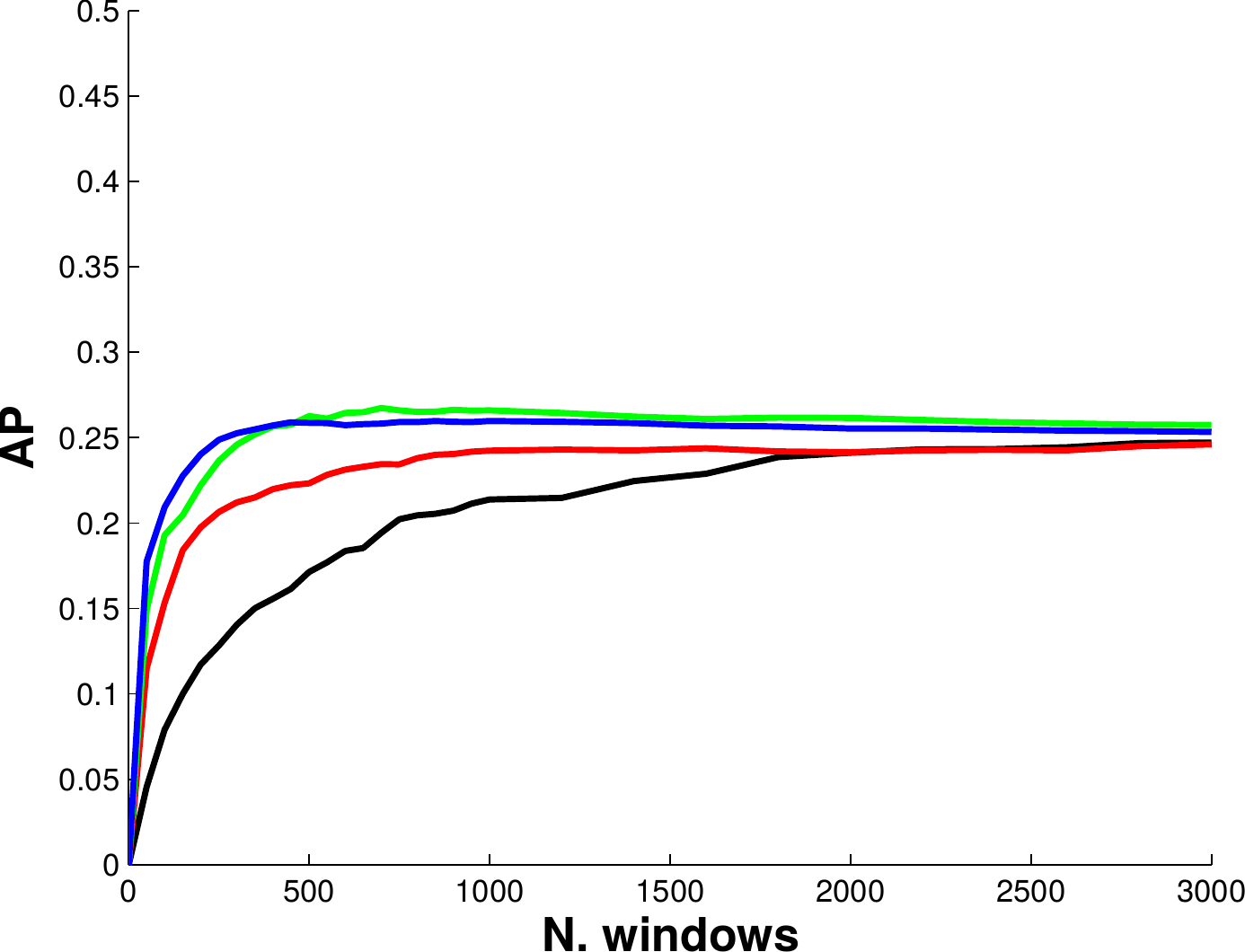}
    \caption{Table}
  \end{subfigure}
~
  \begin{subfigure}[b]{0.3\textwidth}
    \includegraphics[width=\textwidth]{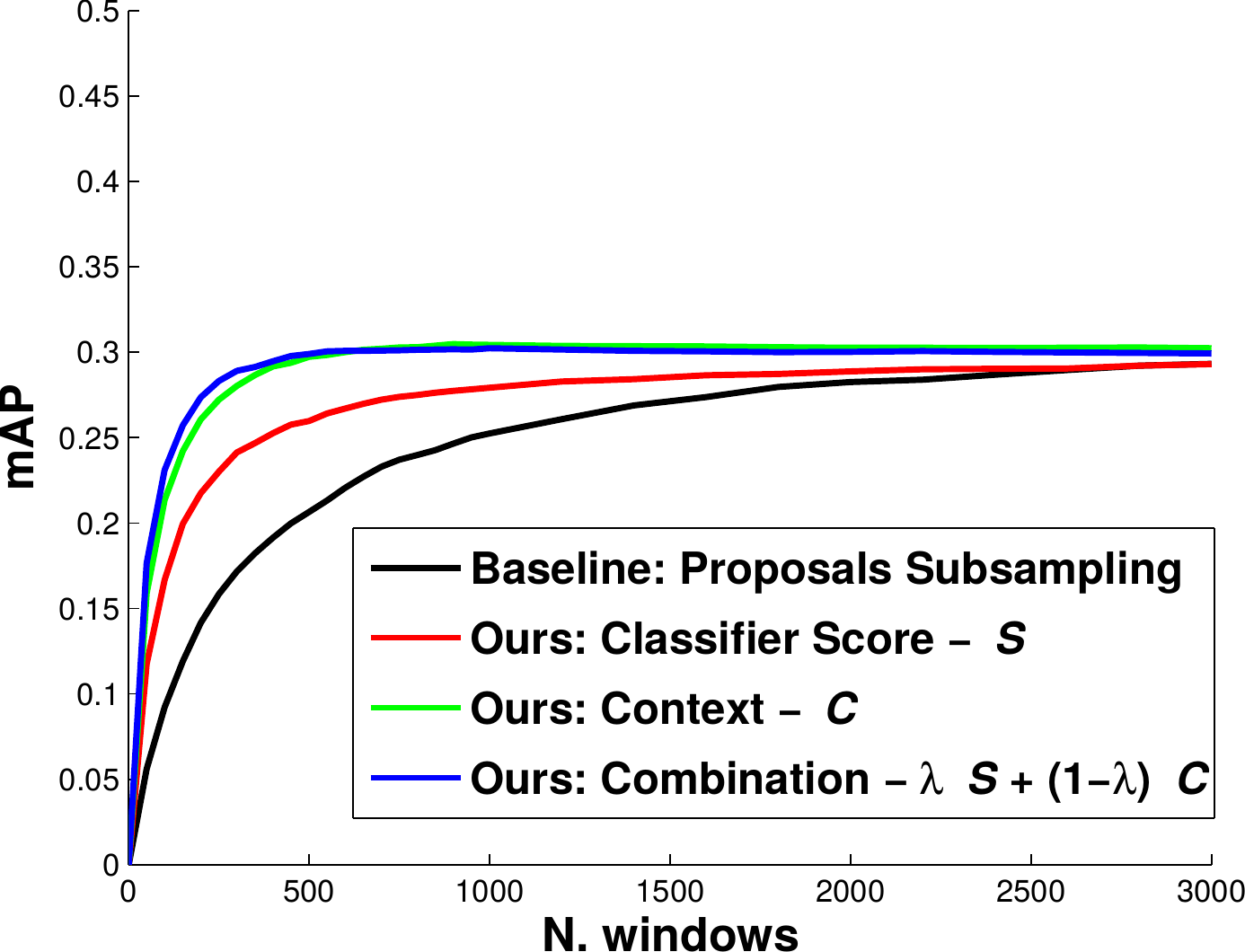}
    \caption{}
  \end{subfigure}

  \end{center}
  \figskiny
  \figsmall
  \caption{\textbf{Results for SUN2012}. \it Results for the baseline Proposals Subsampling and our method on SUN2012, using each force $\mathcal{S}$, $\mathcal{C}$ alone and in combination. 
  The x-axis shows the number of evaluated windows. The y-axis in (a-e) shows the AP of each class,
  while in (f) it shows the mean AP over all classes (mAP).
  \figskiny }
\label{fig:curves}
\end{figure*}

\parskiny
\parskiny
\paragraph{Training hyperparameters $\Theta$.}
For each object class, we find optimal hyperparameters $\sigma_\mathcal{S}$, $\sigma_\mathcal{C}$, and $\lambda$ by maximizing object detection performance by cross-validation on the training set (by grid search in ranges $\sigma_\mathcal{S},\sigma_\mathcal{S}\in[0.01,1]$ and $\lambda\in[0,1]$).
%\todo{should we clarify splits and so on? AG: I don't think so.} 
Performance is quantified by the area under the Average Precision (AP) curve, which reports AP as a function of the number of proposals evaluated by the strategy (fig.~\ref{fig:curves}).
%We train specific $\sigma_\mathcal{S}$ values when only using $\mathcal{S}$, and $\sigma_\mathcal{S}$ when only using $\mathcal{C}$, $\sigma_{\mathcal{C}}$. 
Interestingly, the learned $\sigma$ values correspond to intuition.
For $\sigma_{\mathcal{S}}$ we obtain small values, as the classifier score informs only about the immediate neighborhood of the observed window.
Values for $\sigma_{\mathcal{C}}$ are larger, as the context force informs about a broader region of the image.
Furthermore, $\mathcal{C}$ uses arbitrary windows, hence the distance to proposals is generally larger.

%\todo{AG: Specific $\sigma$ per force, then train $\lambda$ has very similar results. Keep previous (training lambda and sigma jointly).}

\section{Experiments}

We perform experiments on two datasets: SUN2012~\cite{xiao10cvpr} and PASCAL VOC10~\cite{Everingham10}. 

\parshrinky
\paragraph{SUN2012.}
We use all available images for the 5 most frequent classes in the highly challenging SUN2012 dataset~\cite{xiao10cvpr}: Chair, Person, Car, Door and Table. 
This amounts to 2920 training images and 6794 test images, using the official train/test split provided with the dataset~\cite{SUN2012}.
%\todo{really? Or provided by the website? Mention version of the dataset, by date or version number, cite website. This is a tricky thing.}
%\todo{AG: It's provided by their website. They say that if we use the dataset, we should cite that paper and refer to the data as SUN2012. Maybe add URL in footnote?}
Each image is annotated with bounding-boxes on instances of these 5 classes. 
%\todo{really? AG: No, not really. Should we say here that it is not complete, though? Maybe drop the whole sentence?}
This dataset contains large cluttered scenes with small objects, as it was originally created for scene recognition~\cite{xiao10cvpr}. 
This makes it very challenging for object detection, and also well suited to show the benefits of using context in the search strategy.
%To our knowledge, only~\cite{choi10cvpr,Salakhutdinov11cvpr} have used SUN for object detection, and only in its 2009 version.

\parshrinky
%\parskiny
%\figsmall
\paragraph{PASCAL VOC10.}
We use all 20 classes of PASCAL VOC10~\cite{Everingham10}.
While also challenging, on average this dataset has larger and more centered objects than SUN2012.
We use the official splits, train on the \texttt{train} set (4998 images) and test on the \texttt{val} set (5105 images).

\parshrinky
%\parskiny
%\figsmall
\paragraph{Protocol.}
We train the window classifier (including fine-tuning for R-CNN), the Random Forest regressor and the hyperparameters $\Theta$ on the training set.
We measure performance on the test set by Average Precision (AP), following the PASCAL protocol~\cite{Everingham10} (i.e. a detection is correct if it overlaps a ground-truth object $>0.5$).
Previous to the AP computation, we use Non-Maxima Suppression~\cite{felzenszwalb10pami} to remove duplicate detections.

\figsmall
\subsection{Results}
\figsmall
%\begin{table}[t]
%\resizebox{\columnwidth}{!}
%{
%%\footnotesize
%  \small
%\begin{tabular}{|l|r|r|r|r|r|}
%\hline
%\textbf{Class} & \multicolumn{0}{l|}{\textbf{chair}} & \multicolumn{1}{l|}{\textbf{person}} & \multicolumn{1}{l|}{\textbf{car}} & \multicolumn{1}{l|}{\textbf{door}} & \multicolumn{1}{l|}{\textbf{table}} \\ \hline
%\textbf{AP (all)} & 0.301 & 0.316 & 0.451 & 0.176 & 0.253 \\ \hline
%\textbf{Max AP (ours)} & 0.304 & 0.318 & 0.455 & 0.193 & 0.260 \\ \hline
%\textbf{\#Win for max} & 500 & 2000 & 2600 & 500 & 850 \\ \hline
%\end{tabular}
%}
%\figshrinky
%\caption{\textbf{Maximum performance of our method.} \it Comparison between the AP  obtained by evaluating all proposals and the maximum achieved by our method. \figshrinky}
%\label{tab:bestmAP}
%\end{table}

\paragraph{R-CNN on SUN2012.}
Fig.~\ref{fig:curves} presents results for our full system (`Combination') and when using each force $\mathcal{S}$ or $\mathcal{C}$ alone.
The figure shows the evolution of AP as a function of the number of proposals evaluated.
As a baseline, we compare to evaluating proposals in a random sequence (`Proposals Subsampling'). This represents a naive way of reducing the number of evaluations.
The rightmost point on the curves represent the performance of evaluating all proposals, i.e. the original R-CNN method.

Our full method clearly outperforms Proposals Subsampling, by selecting a better sequence of proposals to evaluate.
On average over all classes, by evaluating about 350 proposals we match the performance of evaluating all proposals (fig.~\ref{fig:curves}f).
This corresponds to a $9\times$ reduction in the number of window classifier evaluations.

In general, we achieve our best results by combining both forces $\mathcal{S}$ and $\mathcal{C}$.
When using one force alone, $\mathcal{C}$ performs better, in some cases even reaching the accuracy of our combined strategy. Nevertheless, force $\mathcal{S}$ achieves surprisingly   good results by itself, providing a rather simple method to speed-up state-of-the-art object detectors while maintaining high accuracy.

\begin{figure*}
\begin{center}
  \includegraphics[width=0.9\textwidth]{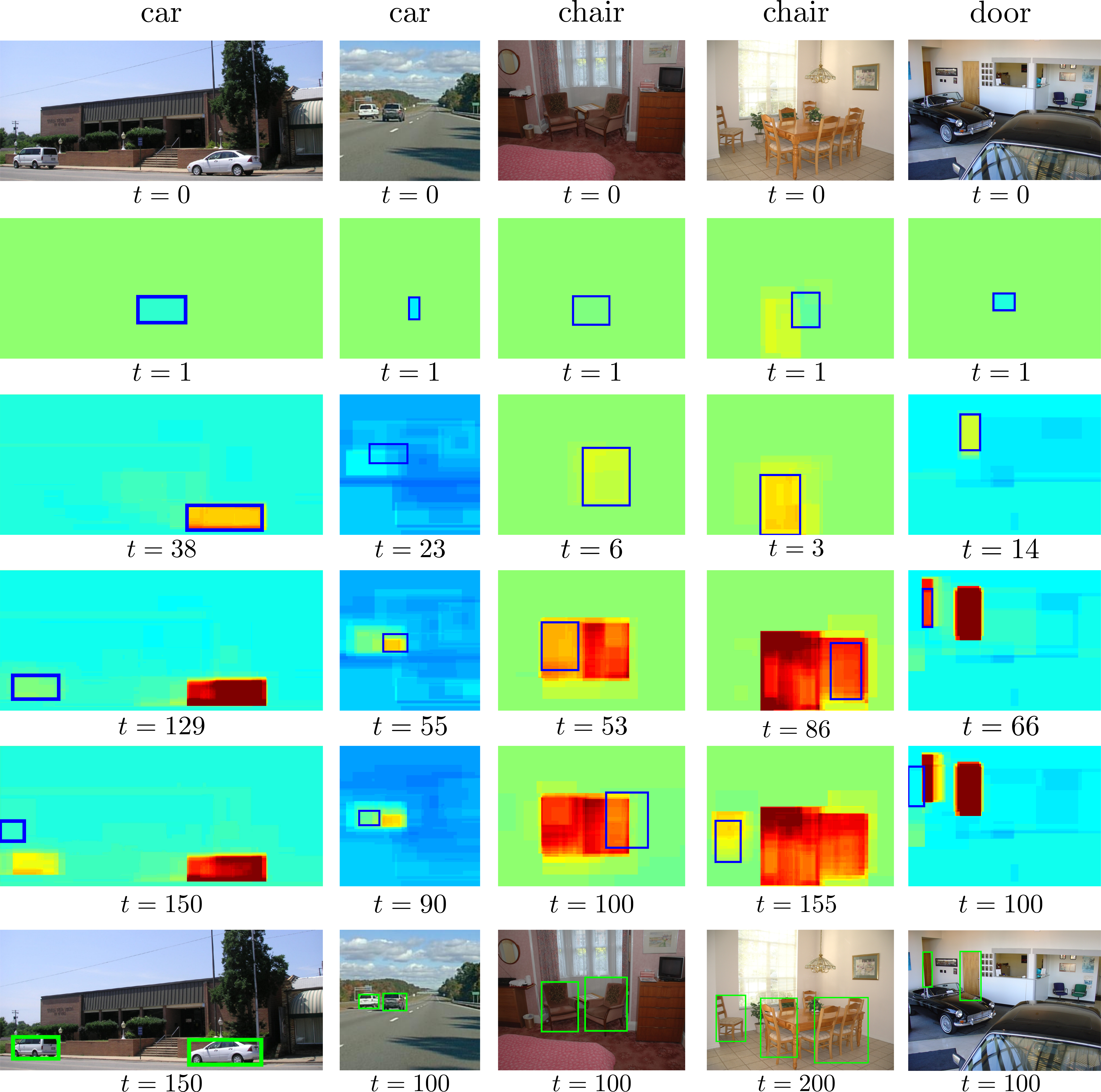}
\end{center}
\figskiny
\caption{\textbf{Qualitative results.} \it 
 (Top) Original image. 
 (Middle) Belief maps for some iterations. The blue window indicates the observation at that iteration.
 (Bottom) The top scored detections. %, where $k$ is the number of ground-truth objects in the image.
 \figskiny \figsmall} 
\label{fig:detectionResults}
\end{figure*}

Fig.~\ref{fig:detectionResults} shows our search strategy in action.
After just a few iterations the belief maps are already highlighting areas containing the objects.
Uninteresting areas such as the sky or ceiling are barely ever visited, hence we waste little computation.
Our method detects multiple object instances and viewpoints, even in challenging images with small objects.
In these examples, it finds the first instance in fewer than 50 iterations, showing its efficiency in guiding the search. After finding the first instance, it continues exploring other areas of the image looking for more instances. 
%In fewer than 300 iterations, the method effectively finds most of the objects in the images.
%\todo{more comments: notice multiple instances, vps, small objects, how quickly homes in to an object, often in the first 20 iterations already.AG: done.}

As a last experiment, we compare our RF context extractor with one based on nearest-neighbour search, as in~\cite{alexe12nips}.
We run our method only using the context force $\mathcal{C}$, but substituting RF with nearest-neighbours, on the same
training set and input features.
The results show that both ways of extracting context lead to the same performance.
The AP at 500 windows, averaged over all classes, differs by only 0.006.
%After 750 iterations, the curves are practically identical.
% We are going to compare our context extractor, Distance RF, with a context extractor using a RF with stumps.
Importantly, however, RF is $60\times$ faster (sec.~\ref{sec:runtime}).

\begin{figure}
  \begin{center}
  \begin{subfigure}[b]{0.23\textwidth}
    \includegraphics[width=\textwidth]{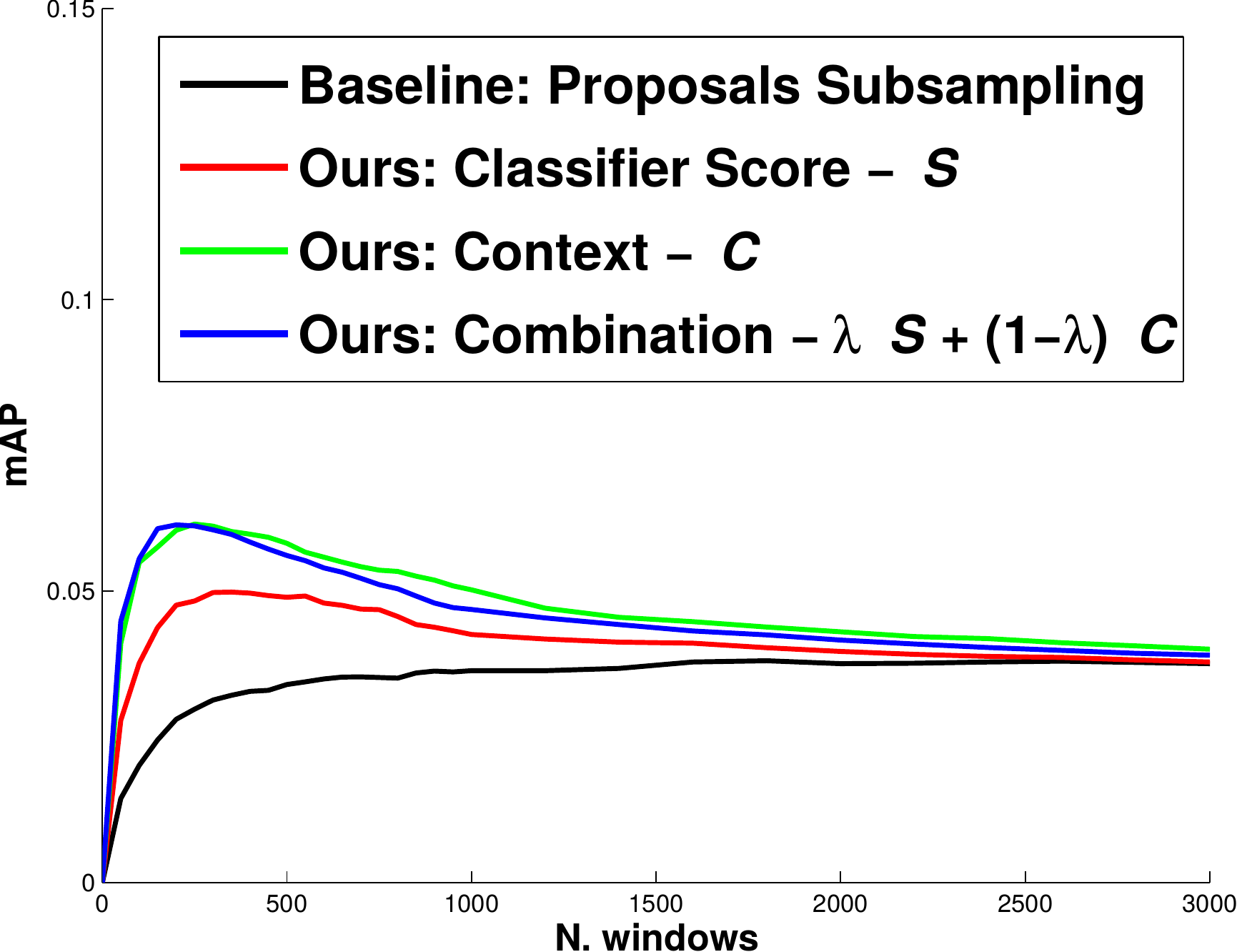}
    \figskiny
  \end{subfigure}
    ~
  \begin{subfigure}[b]{0.23\textwidth}
    \includegraphics[width=\textwidth]{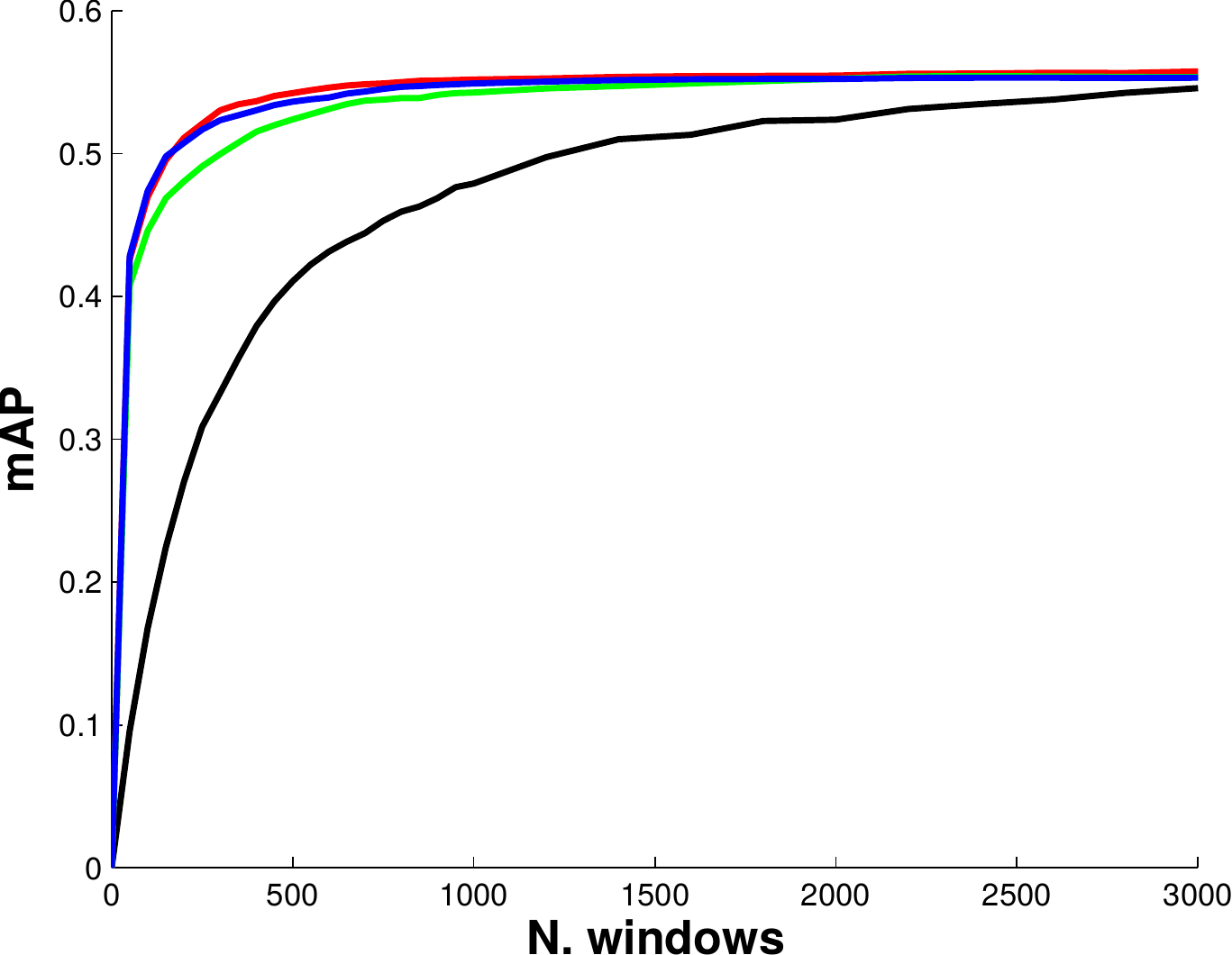}
    \figskiny
  \end{subfigure}

  \end{center}
  \figskiny
  \figsmall
  \caption{\it (Left) Results on SUN2012 using the UvA detector~\cite{uijlings13ijcv}.
  (Right) Results on PASCAL VOC10 using R-CNN.\figskiny \figsmall}
  \label{fig:curvePASCALBoW}
\end{figure}

\parshrinky
\paragraph{UvA on SUN2012.}
We re-trained all the elements of our method on the Bag-of-Words features of~\cite{uijlings13ijcv}: window classifier, RF regressor, and hyperparameters.
Our method matches the performance of evaluating all proposals with 35 windows on average, a reduction of $85\times$ (fig.~\ref{fig:curvePASCALBoW}-left) .
Interestingly, the curve for our method reaches an even {\em higher} point when evaluating just 100 windows (+0.02 mAP).
This is due to avoiding some cluttered areas where the window classifier would produce false-positives.
This effect is less noticeable for the R-CNN window classifier, as UvA is more prone to false-positives.

\parshrinky
\paragraph{R-CNN on PASCAL VOC10.}
%We use CNN features fine-tuned on PASCAL VOC12. \todo{blame girshick}
As fig.~\ref{fig:curvePASCALBoW}-right shows, our method outperforms the Proposal Subsampling baseline again, having a very rapid growth in the first 100 windows.
%It matches mAP of evaluating all proposals at 500 windows.
\parskiny
\figsmall
\label{sec:results}

\seckiny
\subsection{Runtime}
\label{sec:runtime}
\figsmall

We measure runtimes on an Intel Xeon E5-1620v2 CPU and a GeForce GTX 770 GPU, for the R-CNN detector.
The most expensive component is computing CNN features, which takes 4.5 ms per window on the GPU.
%We do not include the object proposal extraction time (about 4 seconds) in any reported timing, as it is performed once per image and it is common to all the methods compared.
Evaluating the 3200 proposals in an average image takes 14.4 seconds.
%\todo{what do you mean 'eval'? The CNN feats, or all of the pipeline? AG: CNN features, normalization and SVM (mostly CNN features, the others take only 0.15s overall). The time used for generating the proposals is not included anywhere.}
The total overhead added by our method is 2.6 ms per iteration.
Therefore, processing one image while maintaining the same AP performance (350 iterations on SUN2012) takes $350 \cdot (4.5 + 2.6)$ ms $= 2.5$ seconds, i.e. $6\times$ faster than evaluating all proposals
\footnote{Extracting CNN descriptors on a GPU is more efficient in batches than one at a time, and is done in R-CNN~\cite{girshick14cvpr}
by batching many proposals in a single image. In our sequential search we can form batches by processing one window each from many different images.}.

The small overhead added by our method is mainly due to the RF query performed at each iteration for the context force, which amounts to 1.9 ms including the Hamming embedding of the appearance features. 
In comparison, a context extractor implemented using nearest-neighbours as in~\cite{alexe12nips} takes 57 ms per iteration, which would lead to no actual total runtime gain over evaluating all proposals.

As the runtime of the window classifier grows, the speedup made by of our method becomes more important. Extracting CNN features on the CPU takes 100 ms per window~\cite{jia13caffe}.
In this regime, the overhead added by our method becomes negligible, only 3\% of
running the window classifier.
Evaluating all 3200 proposals in an image would require 320 seconds,
in contrast to just 36 seconds for evaluating 350 proposals with our method,
a $9\times$ speed-up.

\seckiny
\subsection{Conclusion}
\figsmall
Most object detectors independently evaluate a classifier on all windows in a large set.
Instead, we presented an active search strategy that sequentially chooses the next window to evaluate based on all the information
gathered before.
Our search effectively combines two complementing driving forces: context and window classifier score.
%Given an observation, the context force gives information about the presence of the object in any area of the image.
%We use a novel version of RF, Distance RF, in order to efficiently model the context force.
%On the other hand, the classifier score force informs about the surroundings of the observation.

In experiments on SUN2012 and PASCAL VOC10, our method substantially reduces the number of window classifier evaluations.
Due to the efficiency of our proposed context extractor based on Random Forests, we add little overhead to the detection pipeline, obtaining significant speed-ups in actual runtime.

\parshrinky
{ \paragraph{Acknowledgements}
This work was supported by the ERC Starting Grant VisCul. A. Vezhnevets is also supported by SNSF fellowship PBEZP-2142889.}

{\small
\bibliographystyle{ieee}
\bibliography{../../bibtex/shortstrings,../../bibtex/vggroup,../../bibtex/calvin}
}

\end{document}